\documentclass[10pt,twocolumn,letterpaper]{article}
\usepackage{cvpr}              %

\makeatletter
\@namedef{ver@everyshi.sty}{}
\makeatother
\usepackage{multirow}
\usepackage{graphicx}
\usepackage{amsmath}
\usepackage{amssymb}
\usepackage{booktabs}
\usepackage{inconsolata}
\usepackage{pifont}
\usepackage[labelfont=bf]{caption}
\usepackage[draft]{changes}
\definechangesauthor[name={Roland}, color=blue]{RK}
\usepackage{enumitem}
\usepackage[outercaption]{sidecap}    
\usepackage{empheq}
\usepackage[most]{tcolorbox}
\usepackage{colortbl}
\usepackage{scrextend}
\newcommand{\gradicon}{\texttt{GradICON}} 
\newcommand{\lsim}{\mathcal{L}_{\text{sim}}} 
\newcommand{\lreg}{\mathcal{L}_{\text{reg}}}

\newcommand{\lreggicon}{\mathcal{L}_{\text{reg}}^{\texttt{GradICON}}} 
\newcommand{\lregicon}{\mathcal{L}_{\text{reg}}^{\texttt{ICON}}}

\usepackage[pagebackref,breaklinks,colorlinks,linkcolor=blue,citecolor=blue]{hyperref}

\usepackage[capitalize]{cleveref}
\crefname{section}{Sec.}{Secs.}
\Crefname{section}{Section}{Sections}
\Crefname{table}{Table}{Tables}
\crefname{table}{Tab.}{Tabs.}

\def\cvprPaperID{5950} %
\def\confName{CVPR}
\def\confYear{2023}
\usepackage{amsthm}
\usepackage{marvosym}
\theoremstyle{definition}

\usepackage[utf8]{inputenc} %
\usepackage[T1]{fontenc}    %
\usepackage{amsfonts}       %
\usepackage{nicefrac}       %
\usepackage{microtype}      %
\usepackage{xcolor}         %
\usepackage{multirow}
\usepackage{dsfont}
\usepackage{xfrac}
\usepackage{todonotes}
\usepackage{subcaption}
\usepackage{tablefootnote}
\usepackage{wrapfig}

\begin{document}
\setlength{\abovedisplayskip}{3pt}
\setlength{\belowdisplayskip}{3pt}
\setlength{\abovedisplayshortskip}{3pt}
\setlength{\belowdisplayshortskip}{3pt}

\title{\texttt{GradICON}: Approximate Diffeomorphisms via Gradient Inverse Consistency}
\author{
Lin Tian$^*$\textsuperscript{1}$\;$
Hastings Greer\thanks{Equal Contribution.}$\;$\textsuperscript{1}$\;$
Fran\c{c}ois-Xavier Vialard\textsuperscript{2,3}$\;$
Roland Kwitt\textsuperscript{4}$\;$
Raúl San José Estépar\textsuperscript{5}\\
Richard Jarrett Rushmore\textsuperscript{6}$\;$
Nikolaos Makris\textsuperscript{5}$\;$
Sylvain Bouix\textsuperscript{7}$\;$
Marc Niethammer\textsuperscript{1} \vspace{5pt}\\ 
\textsuperscript{1}UNC Chapel Hill $\;$
\textsuperscript{2}LIGM, Universit\'e Gustave Eiffel$\;$
\textsuperscript{3}MOKAPLAN, INRIA Paris \\
\textsuperscript{4}University of Salzburg $\;$
\textsuperscript{5}Harvard Medical School $\;$
\textsuperscript{6}Boston University $\;$
\textsuperscript{7}\'ETS Montr\'eal
}

\maketitle
\vspace{-0.15cm}
\begin{abstract}
	\vspace{-0.15cm}
	We present an approach to learning regular spatial transformations between image pairs in the context of medical image registration. Contrary to optimization-based registration techniques and many modern learning-based methods, we do not directly penalize transformation irregularities but instead promote transformation regularity via an inverse consistency penalty. We use a neural network to predict a map between a source and a target image as well as the map when swapping the source and target images. Different from existing approaches, we compose these two resulting maps and regularize deviations of the \emph{Jacobian} of this composition from the identity matrix. This regularizer -- \texttt{\textit{GradICON}} -- results in much better convergence when training registration models compared to promoting inverse consistency of the composition of maps directly while retaining the desirable implicit regularization effects of the latter. We achieve state-of-the-art registration performance on a variety of real-world medical image datasets using a single set of hyperparameters and a single non-dataset-specific training protocol. Code is available at \url{https://github.com/uncbiag/ICON}.
	\vspace{-0.8cm}
\end{abstract}

\vspace{-0.15cm}
\section{Introduction}
\vspace{-0.15cm}
\begin{figure}[htp]
	\centering
	\includegraphics[width=\columnwidth]{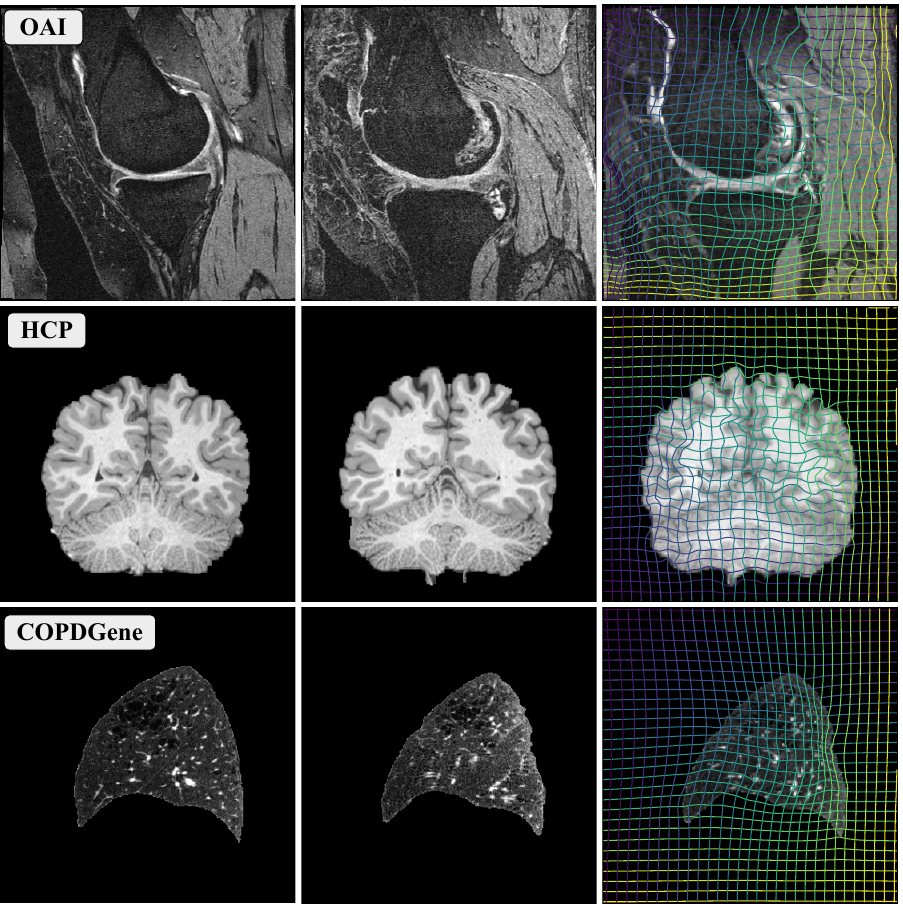}
	\caption{Example source (\emph{left}), target (\emph{middle}) and warped source (\emph{right}) images obtained with our method, trained with a \textbf{single protocol}, using the proposed \texttt{GradICON} regularizer. \label{Fig:teaser}}
	\vspace{-0.8cm}
\end{figure}
Image registration is a key component in medical image analysis to estimate
spatial correspondences between image
pairs~\cite{crum2004non,sotiras2013deformable}. Applications include
estimating organ motion between treatment fractions in radiation
therapy~\cite{kessler2006image,han2021deep}, capturing disease
progression~\cite{viergever2016survey}, or allowing for localized analyses in
a common coordinate system~\cite{evans2012brain}. \vskip0.5ex Many different
registration algorithms have been proposed over the last decades in medical
imaging~\cite{modersitzki2003numerical,miller2006geodesic,vialard2012diffeomorphic,viergever2016survey,chen2021deep}
and in computer vision~\cite{horn1981determining,fortun2015optical}.
Contributions have focused on different transformation models (\ie, what types
of transformations are considered permissible), similarity measures (\ie, how
``good alignment'' between image pairs is quantified), and solution strategies
(\ie, how transformation parameters are numerically estimated). The respective
choices are generally based on application requirements as well as assumptions
about image appearance and the expected transformation space. In consequence,
while reliable registration algorithms have been developed for transformation
models ranging from simple parametric models (\eg, rigid and affine
transformations) to significantly more complex nonparametric
formulations~\cite{modersitzki2003numerical,miller2006geodesic,vialard2012diffeomorphic}
that allow highly localized control, practical applications of registration
typically require many choices and rely on significant parameter tuning to
achieve good performance.
Recent image registration work has shifted the focus from solutions based on numerical optimization for a specific image pair to learning to predict transformations based on large populations of image pairs via neural networks~\cite{yang2017quicksilver,dalca2018unsupervised,chen2021deep,ilg2017flownet,dosovitskiy2015flownet,sun2018pwc,teed2020raft,huang2022flowformer}. However, while numerical optimization is now replaced by training a regression model which can be used to quickly predict transformations at test time, parameter tuning remains a key challenge as loss terms for these two types of approaches are highly related (and frequently the same). Further, one also has additional choices regarding network architectures. Impressive strides have been made in optical flow estimation as witnessed by the excellent performance of recent approaches~\cite{huang2022flowformer} on Sintel~\cite{butler2012naturalistic}. However, our focus is medical image registration, where smooth and often diffeomorphic transformations are desirable; here, a simple-to-use learning-based registration approach, which can adapt to different types of data, has remained elusive. In particular, nonparametric registration approaches require a balance between image similarity and regularization of the transformation to assure good matching at a high level of spatial regularity, as well as \emph{choosing} a suitable regularizer. This difficulty is compounded in a multi-scale approach where registrations at multiple scales are used to avoid poor local solutions.

\vskip0.5ex
Instead of relying on a complex spatial regularizer, the recent \texttt{ICON} approach~\cite{greer2021icon} uses only inverse consistency to regularize the sought-after transformation map, thereby dramatically reducing the number of hyperparameters to tune. While inverse consistency is not a new concept in image registration and has been explored to obtain transformations that are inverses of each other when swapping the source and the target images~\cite{christensen2001consistent}, \texttt{ICON}~\cite{greer2021icon} has demonstrated that a sufficiently strong inverse consistency penalty, by itself, is sufficient for spatial regularity when used with a registration network. Further, as \texttt{ICON} does not explicitly penalize spatial gradients of the deformation field, it does not require pre-registration (\eg, rigid or affine), unlike many other related works. However, while conceptually attractive, \texttt{ICON} suffers from the following limitations: 1) training convergence is slow, rendering models costly to train; and 2) enforcing approximate inverse consistency strictly enough to prevent folds becomes increasingly difficult at higher spatial resolutions, necessitating a suitable schedule for the inverse consistency penalty, which is not required for \texttt{GradICON}.
\vskip0.5ex
Our approach is based on a surprisingly simple, but effective observation: penalizing the Jacobian of the inverse consistency condition instead of inverse consistency directly\footnote{\ie, penalizing deviations from $\nabla(\Phi_\theta^{AB}\circ\Phi_\theta^{BA}-\operatorname{Id})=0$ instead of deviations from $\Phi_\theta^{AB}\circ\Phi_\theta^{BA}-\operatorname{Id}=0$.} applies zero penalty for inverse consistent transform pairs but 1) yields significantly improved convergence, 2) no longer requires careful scheduling of the inverse consistency penalty, 3) results in spatially regular maps, and 4) improves registration accuracy. These benefits facilitate a \emph{unified} training protocol with the same network structure, regularization parameter, and training strategy across registration tasks.
\vskip0.5ex
\noindent
{\bf Our contributions are as follows:}
\vspace{-0.1cm}
\begin{itemize}[leftmargin=*]
	\item We develop \texttt{GradICON} ({\texttt{Grad}}ient {\texttt{I}}nverse
	      {\texttt{CON}}sistency), a versatile regularizer for learning-based image
	      registration that relies on penalizing the Jacobian of the inverse consistency
	      constraint and results, empirically and theoretically, in spatially
	      well-regularized transformation maps. \vspace{-0.1cm}
	\item We demonstrate state-of-the-art (SOTA) performance of models trained with
	      \texttt{GradICON} on three large medical datasets: a knee magnetic resonance
	      image (MRI) dataset of the Osteoarthritis Initiative
	      (OAI)~\cite{nevitt2006osteoarthritis}, the Human Connectome Project's
	      collection of Young Adult brain MRIs (HCP)~\cite{van2012human}, and a computed
	      tomography (CT) inhale/exhale lung dataset from
	      COPDGene~\cite{regan2011genetic}. %
\end{itemize}

\vspace{-0.15cm}
\section{Related work}
\label{sec:related_work}
\vspace{-0.15cm}
\noindent
\textbf{Nonparametric transformation models \& regularization.} There are various ways of modeling a transformation between image pairs. The most straightforward nonparametric approach is via a displacement field~\cite{thirion1998image}. Different regularizers for displacement fields have been proposed~\cite{holden2007review}, but they are typically only appropriate for small displacements~\cite{modersitzki2003numerical} and cannot easily guarantee diffeomorphic transformations~\cite{ashburner2007fast}, which is our focus here for medical image registration. Fluid models, which parameterize a transformation by velocity fields instead, can capture large deformations and, given a suitably strong regularizer, result in diffeomorphic transformations. Popular fluid models are based on viscous fluid flow~\cite{christensen19943d,christensen1996deformable}, the large deformation diffeomorphic metric mapping (LDDMM) model~\cite{beg2005computing}, or its shooting variant~\cite{miller2006geodesic,vialard2012diffeomorphic}. Simpler stationary fluid approaches, such as the stationary velocity field (SVF) approach~\cite{arsigny2009fast,vercauteren2008symmetric}, have also been developed. While diffeomorphic transformations are not always desirable, they are often preferred due to their invertibility, which allows mappings between images to preserve object topologies and prevent foldings that are physically implausible. These models have initially been developed for pair-wise image registration where solutions are determined by numerical optimization, but have since, with minimal modifications, been integrated with neural networks~\cite{yang2017quicksilver,balakrishnan2019voxelmorph,shen2019networks}. In a learning-based formulation, the losses are typically the same as for numerical-optimization approaches, but one no longer directly optimizes over the parameters of the chosen transformation model but instead over the parameters of a neural network which, once trained, can quickly predict the transformation model parameters.

\vskip0.5ex
Fluid registration models are computationally complex as they require solving a fluid equation (either greedily or via direct numerical integration~\cite{miller2001group}, or via scaling and squaring~\cite{arsigny2009fast}), but can guarantee diffeomorphic transformations. In contrast, displacement field models are computationally cheaper but make it more difficult to obtain diffeomorphic transformations. Solution regularity can be obtained for displacement field models by adding appropriate constraints on the Jacobian~\cite{haber2007image}. Alternatively, invertibility can be encouraged by adding inverse consistency losses, either for numerical optimization approaches~\cite{christensen2001consistent} or in the context of registration networks as is the case for \texttt{ICON}~\cite{greer2021icon}. Similar losses have also been used in computer vision to encourage cycle consistencies~\cite{godard2017unsupervised,yin2018geonet,wang2019cycles,bian2022learning} though they are, in general, not focused on spatial regularity. Most relevant to our approach, \texttt{ICON}~\cite{greer2021icon} showed that inverse consistency alone is sufficient to approximately obtain diffeomorphic transformations when the displacement field is predicted by a neural network. \emph{Our work extends this approach by generalizing the inverse consistency loss to a gradient inverse consistency loss, which results in smooth transformations, faster convergence, and more accurate registration results.}

\vskip0.5ex
\noindent
\textbf{Multi-scale image registration.} Finding good solutions for the optimization problems of image registration is challenging, and one might easily get trapped in an unfavorable local minimum. In particular, this might happen for self-similar images, such as lung vessels, where incorrect vessel alignment might be locally optimal. Further, if there is no overlap between vessels, a similarity measure might effectively be blind to misalignment, which is why it is important for a similarity measure to have a sufficient capture range\footnote{Note that keypoint approaches~\cite{hansen2021graphregnet} and approaches based on optimal transport~\cite{shen2021accurate} can overcome some of these issues. However, in this work, we focus on the registration of images with grid-based displacement fields.}.
\vskip0.5ex
Multi-scale approaches have been proposed for optimization-based registration models~\cite{studholme1995multiresolution,maes1999comparative,thevenaz2000optimization,zhao2019recursive} to overcome these issues. For these approaches, the loss function is typically first optimized at a coarse resolution, and the image warped via the coarse transformation then serves as the input for the optimization at a finer resolution. This helps to avoid poor local optima, as solutions computed at coarser resolutions effectively increase capture range and focus on large-scale matching first rather than getting distracted by fine local details. Multi-scale approaches have also been used for learning-based registration~\cite{eppenhof2019progressively,mok2020large,hering2019mlvirnet,jiang2020multi,de2019deep,shen2019networks} and generally achieve better results than methods that only consider one scale~\cite{balakrishnan2019voxelmorph}. These methods all use sub-networks operating at different scales but differ in how the multi-scale strategy is incorporated into the network structure and the training process. A key distinction is if source images are warped as they pass through the different sub-networks~\cite{hering2019mlvirnet,mok2020large,de2019deep,greer2021icon} or if sub-networks always start from the unwarped, albeit downsampled, source image~\cite{eppenhof2019progressively}. The former approach simplifies capturing large deformations as sub-networks only need to refine a transformation rather than capturing it in its entirety.
However, these methods compute the similarity measure and the regularizer losses at all scales, which requires balancing the weights of the losses across all scales and for each scale between the similarity measure and the regularizer. Hence, there are many parameters that are difficult to tune. To side-step the tuning issue, it is common to rely on a progressive training protocol to avoid tuning the weights between losses at all scales. We find that our multi-resolution approach trains well when the loss and regularizer are applied \emph{only} at the highest scale: the coarser components are effectively trained by gradients propagating back through the multi-scale steps.%

\vspace{-0.15cm}
\section{Gradient Inverse Consistency (\texttt{GradICON})}
\label{sec:gradinv}
\vspace{-0.15cm}
\subsection{Preliminaries}
\label{subsec:preliminaries}
\vspace{-0.15cm}
We denote by $I^A:\Omega\,{\to}\,\mathds{R}$ and
$I^B:\,\Omega{\to}\,\mathds{R}$ the \emph{source} and the \emph{target} images in our registration problem. By $\Phi^{AB}: \mathds{R}^d\to\mathds{R}^d$ we denote a \emph{transformation map} with the intention that $I^A \circ \Phi^{AB} \sim I^B$. The map $\Phi^{AB}$ is a \emph{diffeomorphism} if it is differentiable, bijective and its inverse %
is differentiable as well\footnote{Basically, we are interested in properties of $\Phi^{AB}(\Omega)$, as this is the region that can affect the image similarity, but since many maps (\eg, translations) carry points outside of $\Omega$, $\Phi^{BA}$ must be defined at those points for $\Phi^{AB} \circ \Phi^{BA}$ to be defined on all of $\Omega$. In practice, this is achieved for a displacement field $D$ by $\Phi^{AB}:=  x + \texttt{interpolate}(D, \texttt{clip}(x, [0, 1]^d))$.
}.
\emph{Optimization-based} image registration approaches typically solve the optimization problem %
\begin{equation}
	\label{equ:conventional_registration}
	\tau^*=\arg\min_{\tau}\lsim(I^A\circ\varphi^{-1}_{\tau},I^B)+\lambda \lreg(\tau)\enspace,
\end{equation}
where $\lsim(\cdot,\cdot)$ is the \emph{similarity measure}, $\lreg(\cdot)$ is a \emph{regularizer}, $\tau$ are the transformation parameters, and $\lambda\geq 0$. In \emph{learning-based} registration, one does not directly optimize over the transformation parameters of $\varphi^{-1}$, but instead over the parameters $\theta$ of a neural network $\Phi_\theta$ that predicts $\varphi^{-1}$ given the source and target images. Such a network is trained over a set of image pairs $I=\{(I^A_i, I^B_i)\}_{i=1}^N$ by solving

\begin{equation}
\resizebox{0.9 \columnwidth}{!}{
    $\theta^*{=}\arg\underset{\theta}{\min} \frac{1}{N}\sum_{i=1}^N \lsim\left(I^A_i\circ\Phi^{AB}_{\theta,i},I^B_i\right)\,{+}\,
        \lambda\lreg(\Phi^{AB}_{\theta,i})$
}\label{eq:learning_loss}
\end{equation}
with $\Phi^{AB}_{\theta,i}$ as shorthand for  $\Phi_\theta[I^A_i, I^B_i]$ denoting the output of the network given the $i$-th input image pair.  By training with $(I_i^A,I_i^B)$ and $(I_i^B,I_i^A)$ the loss is symmetric in expectation. For ease of notation, we omit the subscripts $i$ or $\theta$ in cases where the dependency is clear from the context.

\vspace{-0.15cm}
\subsection{Regularization}
\label{subsec:approach}
\vspace{-0.15cm}
Picking a good regularizer $\lreg$ is essential as it implicitly expresses the class of transformations one considers plausible for a registration. Ideally, the space of plausible transformations should be known (\eg, based on physical principles) or learned from the data. As nonparametric image registration (at least for image pairs) is an ill-posed problem~\cite{fischer2008ill}, regularization is required to obtain reasonable solutions.
\vskip0.5ex
Regularizers frequently involve spatial derivatives of various orders to discourage spatial non-smoothness~\cite{holden2007review}. This typically requires picking a type of differential operator (or, conversely, a smoothing operator) as well as all its associated parameters. Most often, this regularizer is \emph{chosen} for convenience and not learned from data. Instead of explicitly penalizing spatial non-smoothness, \texttt{ICON}~\cite{greer2021icon} advocates using \emph{inverse consistency} as a regularizer, which amounts to learning a transformation space from data in the class of (approximately) invertible transforms. When implementing inverse consistency, there is a choice of loss. The \texttt{ICON} approach penalizes the sum-of-squares difference between  the identity and the composition of the maps between images $(I^A,I^B)$ and $(I^B,I^A)$, \ie, the regularizer has the form
$ \lregicon = \|\Phi_\theta^{AB}\circ\Phi_\theta^{BA}-\operatorname{Id}\|_2^2\,,$
where $\operatorname{Id}$ denotes the identity transform.
In \cite{greer2021icon}, it is shown that this loss has an \emph{implicit}
regularization effect, similar to a sum-of-squares on the gradient of the
transformation, \ie, an $H^1$ type of norm. In fact, it turns out that regular
invertible maps can be learned without \emph{explicitly} penalizing spatial
gradients. Inspired by this observation, we propose to use the Jacobian
($\nabla$) of the composition of the maps instead, \ie,%
\begin{equation}
	\lreggicon = \left\|\nabla\left[\Phi_\theta^{AB}\circ\Phi_\theta^{BA}\right]-\mathbf{I} \right\|_F^2\enspace,
	\label{eq:grad_icon_regularizer}
\end{equation}
where $\mathbf{I}$ the identity matrix, and $\|\cdot\|_F^2$ is the squared Frobenius norm \emph{integrated} over $\Omega$. As we will see in \cref{sec:experiments}, this loss equally leads to regular maps by exerting another form of implicit regularization, which we analyze in \cref{subsection:analysis}.  To understand the implicit regularization of the \texttt{ICON} loss, one makes the modeling choice $\Phi^{AB}_{\theta} = \Phi^{AB} + \varepsilon n^{AB}$ such that $\Phi^{BA}(\Phi^{AB}) = \operatorname{Id}$, \ie, the output of the network is inverse consistent up to a white noise term $n$ with parameter $\varepsilon >0$ (artificially introduced to make the discussion clear). This white noise can be used to prove that the resulting maps are regularized via the square of a \emph{first-order Sobolev (semi-) norm}. Further, \cite{greer2021icon} empirically showed that an approximate diffeomorphism can be obtained without the white noise when used in the context of learning a neural registration model: if inverse consistency is not exactly enforced,  as suggested above, the inconsistency can be modeled by noise, and the observed smoothness is explained by the theoretical result. In our analysis, we follow a conceptually similar idea.

\vspace{-0.15cm}
\subsection{Analysis}
\label{subsection:analysis}
\vspace{-0.15cm}
\noindent
{\bf Implicit $H^1$ type regularization.}
Since the \texttt{GradICON} loss of \cref{eq:grad_icon_regularizer} is formulated in terms of the gradient, it is a natural assumption to put the white noise on the Jacobians themselves rather than on the maps, \ie, $\nabla \Phi^{AB}_\theta = \nabla \Phi^{AB} + \varepsilon N$ where $N$ is a white noise. This model of randomness is motivated by the stochastic gradient scheme on the \emph{global} population that drives the parameters of the networks. At the level of maps, we write $ \Phi^{AB}_\theta =  \Phi^{AB} + \varepsilon n$ where $N = \nabla n$. Since integration is a low-pass filter, the noise $n$ applies to the low frequencies of $\Phi_\theta^{AB}$. In addition, we expect the low frequencies of the noise to be dampened by the similarity measure between $I^A \circ \Phi^{AB}$ and $I^B$.
Hence, we hypothesize that $\| n \| \ll \| \nabla n \|$, which will be used only once in our analysis. This comparison means that our estimates of the gradient $\nabla n$ and $n$ on our grid satisfy this inequality. We assess this hypothesis \cref{SupplementaryTheoreticalDerivation}.
We start by rewriting the \gradicon~regularizer %
, by applying the chain rule, as
\begin{multline}
	\lreggicon =    \|\left(\nabla \Phi^{AB}(\Phi^{BA}_\theta) + \varepsilon \nabla n^{AB}( \Phi^{BA}_{\theta}) \right) \cdot \\ \left(\nabla \Phi^{BA} + \varepsilon \nabla n^{BA}\right)  - \mathbf{I}\|_F^2\enspace,
	\label{eqn:loss_expanded1}
\end{multline}
using $\nabla \Phi^{AB}_\theta = \nabla \Phi^{AB} + \varepsilon \nabla n^{AB}$
and still omitting the integral sign. We now Taylor expand the loss \wrt $\varepsilon$ and in particular expand the $\nabla \Phi^{AB}(\Phi^{BA}_\theta)$ term from \cref{eqn:loss_expanded1} as
\begin{equation}
	\begin{split}
		& \nabla \Phi^{AB}(\Phi^{BA}_\theta) = \\
		& \nabla \Phi^{AB}(\Phi^{BA}) + \varepsilon \nabla^2 \Phi^{AB}(\Phi^{BA}) n^{BA} + o(\varepsilon)\enspace.
	\end{split}
	\label{eqn:expansion2}
\end{equation}
where $\| n\|~{\ll}~\| \nabla n \|$ implies that the approximation
\begin{equation}
	\nabla \Phi^{AB}(\Phi^{BA}_\theta) = \nabla \Phi^{AB}(\Phi^{BA}) + o(\varepsilon)
\end{equation}
holds as it is only compared with $\nabla n$ in the expansion. Using the first-order approximation $\nabla n^{AB}(\Phi_\theta^{BA}){\approx}\nabla n^{AB}(\Phi^{BA})$, see Appendix \ref{SupplementaryTheoreticalDerivation},
and simplifying Eq. \eqref{eqn:loss_expanded1}, we obtain
\begin{equation}\label{EqExpansion2}
	\resizebox{0.9\columnwidth}{!}{
		$\lreggicon\,{\approx}\,\varepsilon^2 \| \nabla n^{AB}( \Phi^{BA}) \nabla \Phi^{BA}\,{+}\,
			\nabla \Phi^{AB}(\Phi^{BA})  \nabla n^{BA}\|_F^2,$
	}
\end{equation}
as $\nabla \Phi^{AB}(\Phi^{BA})\,{=}\,[\nabla \Phi^{BA}]^{-1}$ (and selecting the first-order coefficients in $\varepsilon$).
Expanding the square then yields
\begin{equation}
    \label{eqn:analysis:squareexpansion}
    \resizebox{.9\columnwidth}{!}{%
    $
    \begin{split}
        & \hspace{-0.8cm}\lreggicon \approx \\
        \hspace{0.15cm}\varepsilon^2
        \Bigl(
        & \left\| \nabla n^{AB}(\Phi^{BA}) \nabla \Phi^{BA}\right\|^2_{F}~+ %
        \left\| \left[\nabla \Phi^{BA}\right]^{-1}\nabla n^{BA} \right\|^2_{F} \\
        & \, + 2 \langle  \nabla n^{AB}(\Phi^{BA}) \nabla \Phi^{BA},\left[\nabla \Phi^{BA}\right]^{-1}\nabla n^{BA} \rangle_{F}
        \Bigr)\,.
    \end{split}
    $
 }
\end{equation}
Under the assumption of independence of the white noises $\nabla n^{AB}$ and $\nabla n^{BA}$, the contribution of the last term in \cref{eqn:analysis:squareexpansion} vanishes in expectation. We are left with the following loss, at order $\varepsilon^2$, now taking expectation,
\vspace{-0.1cm}
\begin{equation}
    \label{EqTakingExpectation}
    \resizebox{.8\columnwidth}{!}{%
    $
    \begin{split}
        \mathbb{E}[\lreggicon] \approx \varepsilon^2 &
        \mathbb{E}
        \Biggl[
        \left\| \nabla n^{AB}(\Phi^{BA}) \nabla \Phi^{BA}\right\|^2_{F}
        ~+ \\
        & \hspace{0.38cm}
        \left\|
        \left[\nabla \Phi^{BA}\right]^{-1}\nabla n^{BA}
        \right\|^2_{F}
        \Biggr] \,.
    \end{split}
     $
 }
\end{equation}
Note that the expectation is explicit due to the white noise assumption; thus, \cref{EqTakingExpectation} can be further simplified to
\begin{equation}
    \label{Equ:regularization_form}
    \resizebox{.8\columnwidth}{!}{%
            $
    	\begin{split}
    		\mathbb{E}[\lreggicon]  \approx \varepsilon^2 &
    		\Biggl(
    		\left\|
    		\left[\nabla \Phi^{AB}\right]^{-1} \sqrt{\operatorname{Det}(\nabla\Phi^{AB})}
    		\right\|_F^2 \\
    		& \hspace{0.23cm}
    		+ \left\|
    		\left[\nabla\Phi^{BA}\right]^{-1}
    		\right\|_F^2\Biggr)\enspace.
    	\end{split}
     $
     }
\vspace{-0.2cm}
\end{equation}
\cref{Equ:regularization_form} amounts to an $L^2$ regularization of the inverse of the Jacobian maps on $\Phi^{AB}$ and $\Phi^{BA}$ which explains why we call it $H^1$ type of regularization, see Appendix \ref{SupplementaryTheoreticalDerivation}; yet, strictly speaking, it is \emph{not} the standard $H^1$ norm \cite{mang2016H1}.
\vskip0.5ex
\noindent
\textbf{Comparison with \texttt{ICON} \normalfont{\cite{greer2021icon}}. }
Interestingly, our analysis shows that \texttt{GradICON} is an $H^1$ type of regularization as for \texttt{ICON}, although we could have expected a second-order regularization from the model. Such higher-order terms appear when taking into account the magnitude of the noise $n$ in the expansion. While there are several assumptions that can be formulated differently, such as the form of the noise and the fact that it is white noise for given pairs of images $(I^A, I^B)$, we believe that \cref{Equ:regularization_form} is a plausible explanation of the observed regularity of the maps in practice. Importantly, since this regularization is implicit, \texttt{GradICON}, as well as \texttt{ICON} can learn based on a slightly more informative prior than this $H^1$ regularization which relies on simplifying assumptions.
\vskip1ex
\noindent
\textbf{Why \texttt{GradICON} performs better than  \texttt{ICON}.}
In practice, we observe that learning a registration model via the \texttt{GradICON} regularizer of \cref{eq:grad_icon_regularizer} shows a faster convergence than using the \texttt{ICON} regularizer, not only in the toy experiment of  Sec.~\ref{sec:exp_convergence_toy_demo} but also on real data. While we do not yet have a clear explanation for this behavior, we provide insight based on the following key differences between the two variants.
\vskip0.5ex
\emph{First}, a difference by a translation is not penalized in the \texttt{GradICON} formulation, but implicitly penalized in the similarity measure, assuming images are not periodic. \emph{Second}, using the Jacobian in \cref{eq:grad_icon_regularizer} correlates the composition of the map between neighboring voxels. In a discrete (periodic) setting, this can be seen by expanding the squared norm. To shorten notations, define $\psi(x) \coloneqq \Phi^{AB}(\Phi^{BA}(x)) - \operatorname{Id}$ considered as a discrete vector. Then, \cref{eq:grad_icon_regularizer} is the sum over voxels  $\sum_{i} \frac{| \psi(x_i+\delta) - \psi(x_i)|^2}{\delta^2}$, which can be rewritten as
\begin{equation}
	\label{eqn:moregradiconcorr}
	\lreggicon = \frac{2 \| \psi \|_{L^2}^2}{\delta^2}   \left(1 - \frac{\langle \psi(\cdot),\psi(\cdot + \delta)\rangle }{\|\psi\|^2_{L^2}}\right)\enspace,
\end{equation}
where $\delta$ is the (hyper-)parameter for the finite difference estimation of the gradient.
This shows that the \texttt{GradICON} regularizer includes more correlation than the \texttt{ICON} regularizer, which would only contain the first factor in \cref{eqn:moregradiconcorr}. \emph{Lastly}, it is known \cite{Charpiat2007} that gradient descent on an $H^1$ energy rather than $L^2$ energy is a preconditioning of the gradient flow, emphasizing high-frequencies over low-frequencies. In our case, however, high-frequencies are the dominant cause of folds. Consequently, as \texttt{GradICON} penalizes high frequencies but allows low frequencies of the composition of the maps to deviate from identity, this is beneficial to avoid folds\footnote{Indeed, the neighborhood of identity of invertible maps is much larger for small-frequency perturbations than for high-frequency perturbations; what matters for invertibility is the norm of the gradient.}.
\emph{Overall, \texttt{GradICON} is more flexible than \texttt{ICON} while retaining the non-folding behavior of the resulting maps}.

\vspace{-0.15cm}
\section{Implementation}
\label{sec:network_architecture}
\vspace{-0.15cm}
To learn a registration model under \cref{eq:grad_icon_regularizer}, we use a neural network that predicts the transformation maps. In particular, we implement a multi-step (\ie, multiple steps within a forward pass), multi-resolution approach trained with a two-stage process, followed by \emph{optional} instance optimization at test time. We will discuss these parts next.

\vspace{-0.15cm}
\subsection{Network structure}
\label{subsection:networkstructure}
\begin{figure}
	\centering
	\includegraphics[width=0.90\columnwidth]{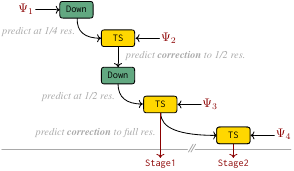}
	\caption{\label{Fig:network_diagram} Illustration of the
		combination steps to create our registration network, see \cref{equ:network_definition}, from the atomic registration networks ($\Psi_i$) via the downsample (\texttt{Down})
		and the two-step (\texttt{TS}) operator.}
	\vspace{-0.4cm}
\end{figure}

\vspace{-0.15cm}
To succinctly describe our network structure, we henceforth omit $\theta$ and represent a \emph{registration neural network} as $\Phi$ (or \eg $\Psi$). The notation $\Phi^{AB}$ (shorthand for $\Phi[I^A, I^B]$) represents the output of this network (a transform from $\mathbb{R}^d \rightarrow{} \mathbb{R}^d$) for input images $I^A$ and $I^B$.  To combine such registration networks into a multistep, multiscale approach, we rely on the following combination operators from~\cite{greer2021icon}:
\begin{equation*}
\resizebox{.9\columnwidth}{!}{%
$
\begin{split}
    & \texttt{Down}\{\Phi\}[I^A, I^B] := \Phi[\texttt{AvgPool}(I^A, 2), \texttt{AvgPool}(I^B, 2)] \\
    & \texttt{TS}\{\Phi, \Psi\}[I^A, I^B] := \Phi[I^A, I^B] \circ \Psi[I^A \circ \Phi[I^A, I^B], I^B]
\end{split}
 $
 }
\end{equation*}
The downsample operator (\texttt{Down}) is for predicting the warp between two high-resolution images using a network that operates on low-resolution images, and the two-step operator (\texttt{TS}) is for predicting the warp between two images in two steps, first capturing the coarse transform via $\Phi$ and then the residual transform via $\Psi$.
We use these operators to realize a multi-resolution, multi-step network, see \cref{Fig:network_diagram}, via
\begin{align}
	\label{equ:network_definition}
	\begin{split}
		&\texttt{Stage1} = \texttt{TS} \{\texttt{Down}\{
		\texttt{TS}\{ \texttt{Down} \{ \Psi_1\}, \Psi_2 \}\}, \Psi_3 \} \,\\
		&\texttt{Stage2}=\texttt{TS} \{ \texttt{Stage1}, \Psi_4\}
	\end{split}
\end{align}
Our \emph{atomic} (\ie, not composite) registration networks $\Psi_i$ are each represented by a UNet instance\footnote{For reference, \texttt{networks.tallUnet2} in the \texttt{ICON} source code.} from \cite{greer2021icon} taking as input two images and returning a \emph{displacement field} $D$. These displacement fields are converted to functions $ x \mapsto x + \texttt{interpolate}(D, x)$ since the above operators are defined on networks that return functions from $\mathbb{R}^d$ to $\mathbb{R}^d$.

\vspace{-0.15cm}
\subsection{Training}
\label{subsection:training}
\vspace{-0.15cm}
We define a \emph{single} training protocol that is applied to train a network on \emph{all} the registration tasks of Sec.~\ref{sec:experiments}. For \emph{preprocessing}, each image has its intensity clipped and rescaled to $[0, 1]$, with clipping intensities appropriate for the modality and anatomy being registered. For modalities with region of interest (ROI) annotations (brain and lung), all values outside the region of interest are set to zero.
For intra-subject registration, we have many fewer pairs to train on\footnote{For intra-subject reg.,
	$N{=}($dataset size$)$ instead of $N{=}(\text{dataset size}^2$).}, so we perform \emph{augmentation} via random flips along axes and small affine warps (see \cref{sec:affine_augmentation}). %
In all experiments, the image similarity measure is combined with the \texttt{GradICON} regularizer to yield the overall loss
\begin{equation}
	\label{equ:loss_function}
	\begin{split}
		\mathcal{L} = & \lsim(I^A \circ \Phi[I^A, I^B], I^B)\, + \\
		& \lsim(I^B \circ \Phi[I^B, I^A], I^A)\, + \\
		&+ \lambda \lVert \nabla (\Phi[I^A, I^B] \circ \Phi[I^B, I^A]) - \mathbf{I}\rVert_F^2\enspace.
	\end{split}
        \vspace{-0.4cm}
\end{equation}
In our implementation, $\nabla (\Phi^{AB} \circ \Phi^{BA})$ is computed using one-sided finite differences with $\Delta x =$1e-3, $\lVert \cdot \lVert^2_F$ is computed by (uniform) random sampling over $\Omega$ with number of samples equal to the number of voxels in the image $/ 2^d$, and we use coordinates where $\Omega = [0, 1]^3$.
\vskip0.5ex
\noindent
\textbf{Multi-stage training.} We train in two stages. $\texttt{Stage1}$: we train the multi-resolution network defined in \cref{equ:network_definition} with the loss from \cref{equ:loss_function}. $\texttt{Stage2}$: we train with the same loss, jointly optimizing the parameters of $\texttt{Stage1}$ and $\Psi_4$.

\vskip0.5ex
\noindent
\textbf{Instance optimization.} We optionally optimize the loss of \cref{equ:loss_function} for 50 iterations at \emph{test time} \cite{wang2022PLOSL}. This typically improves performance but also increases runtime.
\vskip0.5ex
Unless noted otherwise, we use LNCC (local normalized cross-correlation) with a Gaussian kernel (std. of 5 voxels), computed as in~\cite{vishnevskiy2017isotropic}, and \texttt{GradICON} with balancing parameter $\lambda = 1.5$. \texttt{Stage1} and \texttt{Stage2} are trained using ADAM at a learning rate of 5e-5 for 50,000 iterations each. {\bf This protocol remains \emph{fixed} across all datasets}, and any result obtained by exactly this protocol
is marked by $\dagger$; if instance optimization is included, results are marked by $\ddagger$.

\vspace{-0.15cm}
\section{Experiments}\label{sec:experiments}
\vspace{-0.15cm}
\noindent
\textbf{Ethics.} We use one 2D synthetic (\textbf{Triangles and Circles}), one real 2D (\textbf{DRIVE}), and four real 3D datasets (\textbf{OAI}, \textbf{HCP}, \textbf{COPDGene}, \textbf{DirLab}). Acquisitions for all real datasets were approved by the respective Institutional Review Boards.%

\vspace{-0.15cm}
\subsection{Datasets}
\label{subsection:datasets}
\vspace{-0.15cm}
\noindent
\textbf{Triangles and Circles.} A synthetic 2D dataset introduced in \cite{greer2021icon} where images are either triangles or circles. We generate 2000 hollow images with size $128\times128$ which consist of the shape edges and use them as the training set.
\vskip0.5ex
\noindent
\textbf{DRIVE \normalfont{\cite{staal2004ridge}}.} This 2D dataset contains 20 retina
images and the corresponding vessel segmentation masks. We use the segmentation masks to define a synthetic registration problem. In particular, we take downsampled segmentations as the source (/target) image and warp them with random deformations generated by ElasticDeform\footnote{\url{https://zenodo.org/badge/latestdoi/145003699}} to obtain the corresponding target (/source) image. We generate 20 pairs per image, leading to 400 pairs at size $292{\times}282$ in total.
\vskip0.5ex
\noindent
\textbf{OAI \normalfont{\cite{nevitt2006osteoarthritis}}.} We use a subset of 2532 images from the full corpus of MR images of the Osteoarthritis Initiative ({\bf OAI}\footnote{\url{https://nda.nih.gov/oai}}) for training and 301 pairs of images for testing. The images are normalized so that the $1
$th percentile and the $99$th percentile of the intensity are mapped to 0 and 1, respectively. Then we clamp the normalized intensity to be in $[0, 1]$. We follow the train-test split\footnote{Available at \url{https://github.com/uncbiag/ICON}} used in \cite{greer2021icon,shen2019networks}. We downsample images from $160{\times}384{\times}384$ to $80{\times}192{\times}192$
\vskip0.5ex
\noindent
\textbf{HCP \normalfont{\cite{van2012human}}.} We use a subset of T1-weighted and brain-extracted images of size $260{\times}{311}{\times}{260}$ from the Human Connectome Project's (HCP) young adult dataset to assess inter-patient brain registration performance. We downsample images to $130{\times}{155}{\times}{130}$ for training but evaluate the average Dice score for 28 manually segmented subcortical brain regions~\cite{rushmore2022anatomically,rushmore_r_jarrett_2022_dataset}\footnote{\url{https://doi.org/10.5281/zenodo.6967315}} at the original resolution. We train on 1076 images, excluding the 44 images we use for testing.
\vskip0.5ex
\noindent
\textbf{COPDGene \normalfont{\cite{regan2011genetic}}.} We use a subset of
\footnote{The dataset contains 1000 pairs, but 1 pair is also in the \textbf{DirLab} challenge set, and so was excluded from training.} 
999 inspiratory-expiratory lung CT pairs from the {\bf COPDGene}
study\footnote{\url{https://www.ncbi.nlm.nih.gov/gap}}~\cite{regan2011genetic} with provided lung segmentation masks for training. The segmentation masks are computed automatically\footnote{\url{shorturl.at/ciEW6}}. CT images are first resampled to isotropic spacing (2mm) and cropped/padded evenly along each dimension to obtain a $175{\times}{175}{\times}{175}$ volume. Hounsfield units are clamped within $[-1000,0]$ and scaled linearly to $[0,1]$. Then, the lung segmentations are applied to the images to extract the lung region of interest (ROI). Among the processed data, we use 899 pairs for training and 100 pairs for validation.
\vskip0.5ex
\noindent
\textbf{DirLab\normalfont{\cite{castillo2013reference}}.} This dataset is only used to evaluate a trained network. It contains 10 pairs of inspiration-expiration lung CTs with 300 anatomical landmarks per pair, manually identified by an expert in thoracic imaging. We applied the same preprocessing and lung segmentation as for \textbf{COPDGene}.

\vspace{-0.15cm}
\subsection{Ablation study}
\label{subsection:ablationstudy}
\vspace{-0.15cm}
In Table~\ref{table:ablation}, we investigate 1) which UNet structure should be used and 2) how multi-resolution, multi-stage training, instance optimization, and data augmentation affect the registration results. To this end, we train  on \textbf{COPDGene} and evaluate on \textbf{DirLab} using the same similarity measure, regularizer weight $\lambda$, and number of iterations. We report the \emph{mean target registration error (mTRE)} on landmarks (in mm) and the percentage of voxels with negative Jacobian ($\%|J|$).
\newcommand{\abTS}{\phantom{$\dagger$}}
\begin{table}[t!]
	\resizebox{\columnwidth}{!}{%
		\centering
		\begin{tabular}{lcccccc}\toprule
			\textbf{Backbone}                               & \bf Res. & \bf 2nd   & \bf Opt.  & \bf Aug.  & \bf mTRE                & $\%|J|$ \\[-2pt]
			                                                &          &           &           &           & {\footnotesize (in mm)} &         \\[-2pt] \midrule
			UNet from \cite{balakrishnan2019voxelmorph}     & 1        &           &           &           & 17.582\abTS              & 0.01440 \\ \hline %
			\multirow{7}{*}{UNet from \cite{greer2021icon}} & 1        &           &           &           & 5.412\abTS              & 0.00014 \\
			                                                & 3        &           &           &           & 3.259\abTS              & 0.00109 \\
			                                                & 3        &           &           & \ding{51} & 2.587\abTS              & 0.00032 \\
			                                                & 3        & \ding{51} &           &           & 3.182\abTS              & 0.00042 \\
			                                                & 3        & \ding{51} & \ding{51} &           & 1.378\abTS              & 0.00021 \\
			                                                & 3        & \ding{51} &           & \ding{51} & 1.932$\dagger$          & 0.00026 \\
			                                                & 3        & \ding{51} & \ding{51} & \ding{51} & 1.314$\ddagger$         & 0.00023 \\ \bottomrule
		\end{tabular}
	}
	\vspace{-0.1cm}
	\caption{\label{table:ablation}\emph{Ablation results} for training on \textbf{COPDGene} and evaluating on \textbf{Dirlab}. We assess the effect of the backbone network, the number of resolutions (Res.), including \texttt{Stage2} training (2nd), instance optimization (Opt.), and affine augmentation (Aug.).}
	\label{tab:ablation_study}
	\vspace{-0.4cm}
\end{table}
\emph{First}, we observe that the UNet from \cite{greer2021icon}
performs better than the UNet from \cite{balakrishnan2019voxelmorph}. Hence, in all experiments, we adopt the former as our backbone.
Notably, this model has considerably more parameters than the variant from \cite{balakrishnan2019voxelmorph} ($\approx$ 17M \emph{vs.} 300k, see \cref{sec:model_statistics}), but uses less V-RAM as it concentrates parameters in the heavily downsampled layers of the architecture.
\emph{Second}, we find that the multi-resolution approach, including \texttt{Stage2} training, clearly improves performance, reducing the mTRE from 5.412mm to 3.259mm (1 \emph{vs.} 3 res.) and further down to 3.182mm (with \texttt{Stage2}). In this setting, there is also a noticeable benefit of (affine) data augmentation, with the mTRE dropping to 1.932mm. This is not unexpected, though, as we only have 899 pairs of training images for this (inter-patient) registration task. While this is considered a large corpus for medical imaging standards, it is small from the perspective of training a large neural network. \emph{Finally}, we highlight that adding instance optimization yields the overall best results, but the benefits are less noticeable in combination with augmentation.

\vspace{-0.15cm}
\subsection{Comparison to other regularizers}
\label{subsection:regularizercomp}
\vspace{-0.15cm}
We study different regularizers in terms of the trade-off between transform regularity and image similarity on the training set when varying $\lambda$. We use \texttt{Stage1} from \cref{equ:network_definition} for all experiments, setting the regularizer to either \emph{Bending Energy} ($\lreg = \sum_i||\nabla ^2((\Phi^{AB} - \operatorname{Id})_i)||_F^2$), \emph{Diffusion} ($\lreg = || \nabla (\Phi^{AB} - \operatorname{Id})||_F^2$), \texttt{ICON}, or \texttt{GradICON}.
\begin{figure}[htp!]
	\centering
	\includegraphics[width=0.95\columnwidth]{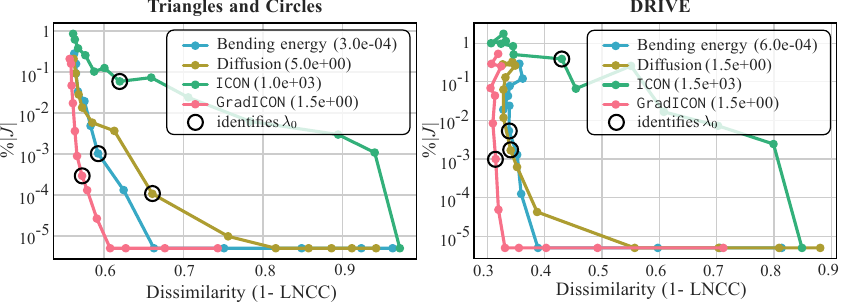}
	\caption{\gradicon~\emph{vs.} other regularization techniques.
		\label{Fig:exp_comp_regularizer_with_varying_lambda}}
	\vspace{-0.25cm}
\end{figure}
Specifically, we pick $\lambda_0$ for each regularizer such that $\%|J|$ is kept at roughly the same level and train multiple networks with $\lambda=\{\lambda_0\cdot 2^i\,|\, i \in [-6,6], i\in \mathbb{Z}\}$ for the same number of iterations. Fig.~\ref{Fig:exp_comp_regularizer_with_varying_lambda} shows results on {\bf Triangles and Circles} and {\bf DRIVE}. We observe that all the regularizers lose a certain level of accuracy when increasing $\lambda$ and \gradicon~in general has the least sacrifice in similarity. This is presumably because of the possible magnitude of deformation each regularizer allows. More results can be found in
\cref{sec:appendix_comparing_to_other_regularizers}.

\vspace{-0.15cm}
\subsection{Empirical convergence analysis} \label{sec:exp_convergence_toy_demo}
\begin{figure}[htp]
	\centering
	\includegraphics[width=0.95\columnwidth]{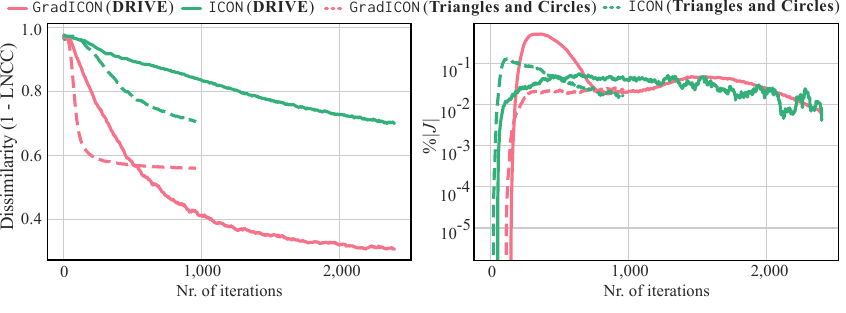}
	\caption{Comparison of the convergence speed (\emph{left}), visualized as $1$-LNCC (\ie, dissimilarity), for \texttt{ICON} and \gradicon~when $\lambda$ is set to produce a similar level of map regularity (\emph{right}).
		\label{Fig:exp_convergence_experiment}}
	\vspace{-0.25cm}
\end{figure}

\vspace{-0.15cm}
To demonstrate improved convergence when training models with our
\texttt{GradICON} regularizer \emph{vs.} models trained with the \texttt{ICON}
regularizer of \cite{greer2021icon}, we assess the corresponding \emph{loss
	curves} under the same network architecture, in particular, the network
described in Sec.~\ref{sec:network_architecture}. We are specifically
interested in (training) convergence behavior when both models produce a
\emph{similar level of map regularity}. To this end, we choose $\lambda$ to
approximately achieve the same similarity loss under both regularizers and
plot the corresponding curves for $\%|J|$, see
\cref{Fig:exp_convergence_experiment}, for \textbf{Triangles and Circles} and
\textbf{DRIVE}.

Overall, we see that \texttt{GradICON} converges significantly faster than
\texttt{ICON}. We hypothesize that this is due to the fact that maps produced
by \texttt{GradICON} contain larger motions than maps produced by
\texttt{ICON}. \Ie, for the same level of regularity, \texttt{ICON} more
strongly limits deformations, effectively slowing down and resulting in less
accurate image alignments.

\vspace{-0.15cm}
\subsection{Inter-patient registration}
\label{subsection:interpatientreg}
\vspace{-0.15cm}
We evaluate inter-patient registration performance of our model with \texttt{GradICON} regularization on \textbf{OAI} and \textbf{HCP}. We report $|\%J|$ (as in \cref{subsection:ablationstudy}) and the mean DICE score between warped and target image for the segmentations of the femoral and tibial cartilage (\textbf{OAI}), and of a set of 28 subcortical brain regions (\textbf{HCP}).
Both measures, averaged over the evaluation data, are listed in  Table~\ref{tab:exp_oai}. In particular, we compare \texttt{GradICON} to the methods reported in ~\cite{shen2019networks} and \cite{greer2021icon} on \textbf{OAI}, and compare to ANTs SyN~\cite{avants2008symmetric} and SynthMorph~\cite{hoffmann2022synthmorph}\footnote{Using SynthMorph networks trained on \emph{HCP aging} data, which differs slightly from the \emph{HCP Young Adults} data we use; see~\cite{harms2018972} for a comparison.} (\texttt{sm-shapes/brains}) on \textbf{HCP}. Segmentations are not used during training and allow quantifying if the network yields semantically meaningful registrations.
Table~\ref{tab:exp_oai} shows that on the \textbf{OAI} dataset \texttt{GradICON} can significantly outperform the state-of-the-art with minimal parameter tuning and, just as \texttt{ICON}, \emph{without} the need for affine pre-alignment (\ie, a step needed by many registration methods on \textbf{OAI}). On \textbf{HCP}, \texttt{GradICON} performs better than the standard
non-learning approach (SyN) and matches SynthMorph while not requiring affine pre-alignment and producing much less folds.
\addtolength{\tabcolsep}{-4.5pt}
\newcommand{\resTS}{\phantom{$\ddagger$}}
\begin{table}[htp]
	\begin{small}
		\centering
		\resizebox{\columnwidth}{!}{
			\begin{tabular}{lccccc}
				\toprule
				\bf Method                                                & \bf Trans.           & $\lreg$                               & $\lsim$ & \bf DICE\, $\uparrow$       & $\%|J|\downarrow$       \\
				\midrule
				\multicolumn{6}{c}{\cellcolor{blue!10}\bf OAI}                                                                                                                                             \\
				Initial                                                   &                      &                                       &         & 7.6\resTS                                             \\
				\midrule
				Demons~\cite{vercauteren2009diffeomorphic}                & A,DVF                & Gaussian                              & MSE     & 63.5\resTS                  & 0.0006                  \\ %
				SyN~\cite{avants2008symmetric}                            & A,VF                 & Gaussian                              & LNCC    & 65.7\resTS                  & 0.0000                  \\
				NiftyReg~\cite{modat2010fast}                             & A,B-Spline           & BE                                    & NMI     & 59.7\resTS                  & 0.0000                  \\
				NiftyReg~\cite{modat2010fast}                             & A,B-Spline           & BE                                    & LNCC    & {\bf 67.9}\resTS            & 0.0068                  \\ %
				vSVF-opt~\cite{shen2019networks}                          & A,vSVF               & m-Gauss                               & LNCC    & 67.4\resTS                  & 0.0000                  \\ \hline
				VM~\cite{balakrishnan2019voxelmorph}                      & SVF                  & Diff.                                 & MSE     & 46.1\resTS                  & 0.0028                  \\ %
				VM~\cite{balakrishnan2019voxelmorph}                      & A,SVF                & Diff.                                 & MSE     & 66.1\resTS                  & 0.0013                  \\ %
				AVSM~\cite{shen2019networks}                              & A,vSVF               & m-Gauss                               & LNCC    & 68.4\resTS                  & 0.0005                  \\ %
				\texttt{ICON}*~\cite{greer2021icon}                       & DVF                  & \texttt{ICON}                         & MSE     & 65.1\resTS                  & 0.0040                  \\ %
				{\textbf{Ours} (MSE, $\lambda{=}0.2$)}                    & DVF                  & \cellcolor{black!10}\texttt{GradICON} & MSE     & 69.5\resTS                  & 0.0000                  \\ %
				{\textbf{Ours} (MSE, $\lambda{=}0.2$, Opt.)}              & DVF                  & \cellcolor{black!10}\texttt{GradICON} & MSE     & 70.5\resTS                  & 0.0001                  \\ %
				\multirow{2}{*}{\textbf{Ours} \emph{(std. protocol)}}     & DVF                  & \cellcolor{black!10}\texttt{GradICON} & LNCC    & 70.1$\dagger$               & 0.0261                  \\ %
				                                                          & DVF                  & \cellcolor{black!10}\texttt{GradICON} & LNCC    & {\bf 71.2}$\ddagger$        & 0.0042                  \\
				\midrule
				\multicolumn{6}{c}{\cellcolor{blue!10}\bf HCP}                                                                                                                                             \\
				Initial                                                   &                      &                                       &         & 53.4\resTS                 &                         \\
				\midrule
				FreeSurfer-Affine$^\ast$~\cite{reuter2010highly}          & A                    & \textemdash                           & TB      & 62.1\resTS                 & 0.0000                  \\
				SyN$^\ast$~\cite{avants2008symmetric}                     & A,VF                 & Gaussian                              & MI      & {\bf 75.8}\resTS            & 0.0000                  \\ \hline
				sm-shapes$^\ast$~\cite{hoffmann2022synthmorph}            & A,SVF                & Diff.                                 & DICE    & 79.8\resTS                  & 0.2981                  \\
				sm-brains$^\ast$~\cite{hoffmann2022synthmorph}            & A,SVF                & Diff.                                 & DICE    & 78.4\resTS                 & 0.0364                  \\
				\multirow{2}{*}{\textbf{Ours} \emph{(std. protocol)}}     & DVF                  & \cellcolor{black!10}\texttt{GradICON} & LNCC    & 78.7$\dagger$               & 0.0012                  \\
				                                                          & DVF                  & \cellcolor{black!10}\texttt{GradICON} & LNCC    & \bf{80.5}$\ddagger$         & 0.0004                  \\
				\midrule
				\multicolumn{6}{c}{\cellcolor{blue!10}\bf DirLab}                                                                                                                                          \\ \midrule
				\bf Method                                                & \bf Trans.           & $\lreg$                               & $\lsim$ & \textbf{mTRE} $\downarrow$  & $\%|J|\downarrow$       \\[-2pt]
				                                                          &                      &                                       &         & {\footnotesize (in mm)}     &                         \\[-2pt]
				Initial                                                   &                      &                                       &         & 23.36\resTS                 &                         \\ \midrule
				SyN~\cite{avants2008symmetric}                            & A,VF                 & Gaussian                              & LNCC    & 1.79\resTS                  & \textemdash             \\
				Elastix~\cite{klein2009elastix}                           & A,B-Spline           & BE                                    & MSE     & 1.32\resTS                  & \textemdash             \\
				NiftyReg~\cite{modat2010fast}                             & A,B-Spline           & BE                                    & MI      & 2.19\resTS                  & \textemdash             \\
				PTVReg~\cite{vishnevskiy2017isotropic}                    & DVF                  & TV                                    & LNCC    & 0.96\resTS                  & \textemdash             \\
				RRN~\cite{he2021recursive}                                & DVF                  & TV                                    & LNCC    & \textbf{0.83}\resTS         & \textemdash             \\
				\midrule
				VM$^\ast$~\cite{balakrishnan2019voxelmorph}               & A,SVF                & Diff.                                 & NCC     & 9.88\resTS                  & 0.0000                  \\
				LapIRN$^\ast$~\cite{mok2020large}                         & SVF                  & Diff.                                 & NCC     & 2.92\resTS                  & 0.0000                  \\
				LapIRN$^\ast$~\cite{mok2020large}                         & DVF                  & Diff.                                 & NCC     & 4.24\resTS                  & 0.0105                  \\
				\multirow{3}{*}{Hering et al. ~\cite{hering2021cnn}}      & \multirow{3}{*}{DVF} & \multirow{3}{*}{Curv+VCC}             & DICE    & \multirow{3}{*}{2.00\resTS} & \multirow{3}{*}{0.0600} \\
				                                                          &                      &                                       & +KP     &                             &                         \\
				                                                          &                      &                                       & +NGF    &                             &                         \\
				GraphRegNet~\cite{hansen2021graphregnet}                  & DV                   & \textemdash                           & MSE     & 1.34\resTS                  & \textemdash             \\
				\multirow{2}{*}{PLOSL~\cite{wang2022PLOSL}}               & \multirow{2}{*}{DVF} & \multirow{2}{*}{Diff.}                & TVD     & \multirow{2}{*}{3.84\resTS} & \multirow{2}{*}{0.0000} \\
				                                                          &                      &                                       & +VMD    &                             &                         \\
				\multirow{2}{*}{$\text{PLOSL}_{50}$~\cite{wang2022PLOSL}} & \multirow{2}{*}{DVF} & \multirow{2}{*}{Diff.}                & TVD     & \multirow{2}{*}{1.53\resTS} & \multirow{2}{*}{0.0000} \\
				                                                          &                      &                                       & +VMD    &                             &                         \\
				\texttt{ICON}$^\ast$~\cite{greer2021icon}                 & DVF                  & \texttt{ICON}                         & LNCC    & 7.04\resTS                  & 0.3792                  \\
				\multirow{2}{*}{\textbf{Ours} \emph{(std. protocol)}}     & DVF                  & \cellcolor{black!10}\texttt{GradICON} & LNCC    & 1.93$\dagger$               & 0.0003                 \\
				                                                          & DVF                  & \cellcolor{black!10}\texttt{GradICON} & LNCC    & \textbf{1.31}$\ddagger$     & 0.0002                  \\
				\bottomrule\vspace{-7mm}
			\end{tabular}}
		\caption{Results on \textbf{OAI}, \textbf{HCP} and \textbf{DirLab}. $\dagger$ and $\ddagger$ indicate results obtained using our standard training protocol (\cref{subsection:training}), w/o ($\dagger$) and w/ ($\ddagger$) instance optimization (Opt.). Only when \texttt{GradICON} is trained with MSE do we set $\lambda=0.2$. Results marked with $^\ast$ are obtained using the official source code; otherwise, values are taken from the literature (see \ref{sec:appendix_sota_comp}). \emph{Top} and \emph{bottom} table parts denote non-learning and learning-based methods, resp. For \textbf{DirLab}, results are shown in the common \emph{inspiration$\rightarrow$expiration} direction.  \underline{A}: affine pre-registration, \underline{BE}: bending energy, \underline{MI}: mutual information, \underline{TB}: Tukey's biweight, \underline{DV}: displacement vector of sparse key points, \underline{TV}: total variation, \underline{Curv}: curvature regularizer, \underline{VCC}: volume change control, \underline{NGF}: normalized gradient flow, \underline{TVD}: sum of squared tissue volume difference, \underline{VMD}: sum of squared vesselness measure difference, \underline{Diff}: diffusion, \underline{VF}: velocity field, \underline{SVF}: stationary VF, \underline{DVF}: displacement vector field. $\underline{\text{PLOSL}_{50}}$: 50 iterations of instance optimization with PLOSL.
		}
		\label{tab:exp_oai}
	\end{small}
\end{table}
\addtolength{\tabcolsep}{4.5pt}

\vspace{-0.15cm}
\subsection{Intra-patient registration}
\label{sec:exp_lung}
\vspace{-0.15cm}
We demonstrate the ability of our model with \texttt{GradICON} regularization to predict \emph{large deformations} between lung exhale (source)/inhale (target) pairs (within patient) from \textbf{COPDGene}. %
This dataset is challenging as deformations are complex and large. Motion is primarily visually represented by the deformation of lung vessels, which form a complex tree-like structure that creates capture-range and local minima challenges for registration. As in our ablation study
(\cref{subsection:ablationstudy}), we report $\%|J|$ and the mTRE (in mm) for manually annotated lung vessel landmarks~\cite{castillo2013reference} averaged over all 10 \textbf{DirLab} image pairs. %
We assess \texttt{GradICON} against traditional optimization-based methods and state-of-the-art (SOTA) learning-based methods. Among the learning-based methods, VM~\cite{balakrishnan2019voxelmorph}, LapIRN~\cite{mok2020large}, Hering el al.~\cite{hering2021cnn} and PLOSL~\cite{wang2022PLOSL} are trained on a large dataset and evaluated on the unseen test dataset. GraphRegNet~\cite{hansen2021graphregnet} has been evaluated on a small dataset with 5-fold cross validation. $\text{PLOSL}_{50}$~\cite{wang2022PLOSL} is PLOSL with instance optimization, which is comparable to \texttt{GradICON}$\ddagger$. Table~\ref{tab:exp_oai} shows that our performance (1.93mm) in a single forward pass is the best of any one-forward-pass neural method that has been trained on a large dataset and is evaluated on an unseen test set. our approach \emph{with} instance optimization (1.31mm) exceeds the SOTA for learning-based approaches of any sort and performs slightly worse than the best optimization-based methods. %

\vspace{-0.15cm}
\section{Conclusion}
\label{section:conclusion}
\vspace{-0.15cm}
We introduced and theoretically analyzed \texttt{GradICON}, a new regularizer to train deep image registration networks. In contrast to \texttt{ICON}~\cite{greer2021icon}, \texttt{GradICON} penalizes the \emph{Jacobian} of the inverse consistency constraint. This has profound effects: we obtain dramatically faster training convergence, higher registration accuracy, do not require scale-dependent regularizer tuning, and retain desirable implicit regularization effects resulting in approximately diffeomorphic transformations. Remarkably, this allows us to train registration networks using \texttt{GradICON} regularization with \emph{one} standard training protocol for a range of different registration tasks. In fact, using this standard training protocol, we match or outperform state-of-the-art registration methods on three challenging and diverse 3D datasets. This uniformly good performance without the need for dataset-specific tuning takes the pain out of training deep 3D registration networks and makes our approach highly practical.

\vskip0.5ex
\noindent
{\bf Limitations and future work.} We only explored first-order derivatives of the inverse-consistency constraint and intensity-based registration. This might have limited registration performance. Studying higher-order derivatives, more powerful image similarity measures (\eg, based on deep features), as well as extensions to piecewise diffeomorphic transformations would be interesting future work.

\vspace{-0.15cm}
\section{Acknowledgements}
\vspace{-0.15cm}
This work was supported by NIH grants 1R01AR072013, 1R01HL149877, 1R01EB028283, RF1MH126732, R41MH118845, R01MH112748, R01NS125307, and 5R21LM013670. The work expresses the views of the authors, not of NIH. Roland Kwitt was supported in part by the Austrian Science Fund (FWF): project FWF P31799-N38 and the Land Salzburg (WISS 2025) under project numbers 20102- F1901166-KZP and 20204-WISS/225/197-2019. The knee imaging data were obtained from the controlled access datasets distributed from the Osteoarthritis Initiative (OAI), a data repository housed within the NIMH Data Archive. OAI is a collaborative informatics system created by NIMH and NIAMS to provide a worldwide resource for biomarker identification, scientific investigation and OA drug development. Dataset identifier: NIMH Data Archive Collection ID: 2343. The brain imaging data were provided by the Human Connectome Project, WU-Minn Consortium (Principal Investigators: David Van Essen and Kamil Ugurbil; 1U54MH091657) funded by the 16 NIH Institutes and Centers that support the NIH Blueprint for Neuroscience Research; and by the McDonnell Center for Systems Neuroscience at Washington University. The lung imaging data was provided by the COPDGene study.
\clearpage
{\small
	\bibliographystyle{ieee_fullname}
	\bibliography{references}
}

\newpage
\appendix
\clearpage
\section*{Overview}
The different sections of the \textbf{supplementary material} cover the following aspects of our \texttt{GradICON} approach.
\begin{itemize}
	\item {\bf Appendix \ref{SupplementaryTheoreticalDerivation}} provides a justification of our noise assumptions, details of the derivation of the regularization properties of \texttt{GradICON}, and additional insights on the convergence behavior of \texttt{GradICON}.
	\item {\bf Appendix \ref{sec:affine_augmentation}} describes our affine data augmentation strategy in detail.
	\item {\bf Appendix \ref{sec:appendix_comparing_to_other_regularizers}} provides detailed comparisons of \texttt{GradICON} to other regularizers, including the loss curve and examples associated with the experiment in Sec.~\ref{subsection:regularizercomp}.
	\item {\bf Appendix \ref{sec:appendix_convergence_experiment}} describes the details of the experiment in Sec.~\ref{sec:exp_convergence_toy_demo} with convergence speed demonstration on \textbf{OAI} dataset.
	\item {\bf Appendix \ref{sec:appendix_sota_comp}} shows an expanded version of Table~\ref{tab:exp_oai} from the main manuscript. In particular, this table provides information on the provenance of these validation results.
	\item {\bf Appendix \ref{sec:model_statistics}} provides details on inference times, memory use, and numbers of parameters for some key learning-based registration approaches.
	\item {\bf Appendix \ref{appendix:visualizations}} shows example registration results for the \textbf{OAI}, \textbf{HCP}, and \textbf{COPDGene} datasets.
	\item {\bf Appendix \ref{Sec:social_impact}} discusses potentials for negative societal impacts of our work.
\end{itemize}
\section{Supplementary material}
\subsection{Analysis details}
\label{SupplementaryTheoreticalDerivation}
\begin{figure}[htp]
	\centering
	\includegraphics[width=1.\linewidth]{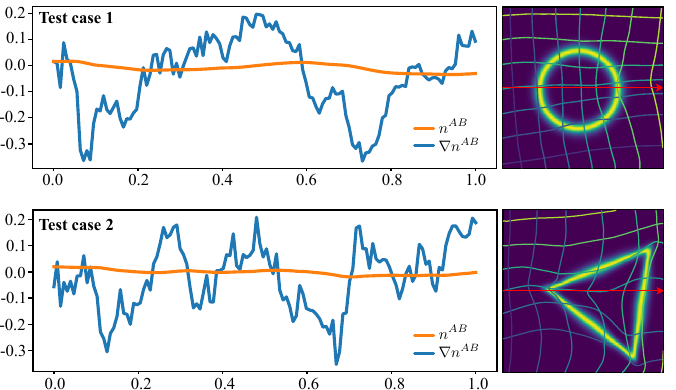}
	\caption{Two \emph{finite difference estimations} of the noise $n^{AB}$ and the gradients $\nabla n^{AB}$ on the synthetic dataset. The magnitude of the gradient is an order of magnitude higher which confirms our hypothesis.}
	\label{fig:NoisesComparisons}
\end{figure}

\noindent
\textbf{Experiments on the main modeling hypothesis.} The main modeling hypothesis in the implicit regularization analysis of \cref{subsection:analysis}
is that the noise term $n$ can be neglected in the Taylor expansion
$\nabla \Phi^{AB}(\Phi^{BA}_\theta) = \nabla \Phi^{AB}(\Phi^{BA}) + \varepsilon \nabla^2 \Phi^{AB}(\Phi^{BA})( n^{BA}) + o(\varepsilon)$.
In this formula, we argue in the main text that the noise term $n^{BA}$ can be neglected  with respect to $\varepsilon$, which is the scale of the noise on the Jacobian. Indeed, only the low-frequency noise should appear since integration is a low-pass filter, but we expect this low-frequency noise to be dampened by the similarity measure, which is an $L^2$ norm on the images. On the synthetic dataset, we checked that our hypothesis is valid as a first approximation. From a given output of the network $\Phi^{AB}_\theta,\Phi^{BA}_\theta$, we estimated the closest $\Phi^{AB},\Phi^{BA}$ in $L^2$ norm to our data. Although this estimate is certainly biased, it is the first natural estimator to check our assumption. In \cref{fig:NoisesComparisons}, we plot the noise $n^{AB}$ and its corresponding gradient $\nabla n^{AB}$ in one chosen \emph{direction} (indicated by the red arrow in the plots on the right-hand side).
\vskip1ex
\noindent
\textbf{Derivation details.}
Some steps that were omitted in the main text are explained hereafter.
Our main object of interest is the \texttt{GradICON} regularizer
\begin{equation}
	\lreggicon = \left\|\nabla_x [\Phi_\theta^{AB} (\Phi_\theta^{BA}(x))] - \mathbf{I}\right\|_F^2\enspace.
	\label{app:eqn:gradicon}
\end{equation}
In what follows, we use $\nabla$ insteand of $\nabla_x$.
Making the assumption that the neural network will try to be perfectly inverse consistent in order to minimize \cref{app:eqn:gradicon}, but will make some error due to limited capacity or imprecision in the training process or just because of trying to balance the matching term and the inverse consistency, we decompose each neural network output $\Phi_\theta^{AB}, \Phi_\theta^{BA}$ into two components, a perfectly inverse consistent component $\Phi^{AB}$, and random noise $\varepsilon n^{AB}$, \ie,
\begin{equation}
	\lreggicon = \left\| \nabla[\Phi^{AB} (\Phi_\theta^{BA}) + \varepsilon n^{AB}(\Phi_\theta^{BA})] -  \mathbf{I} \right\|^2_F\enspace.
\end{equation}
By applying the chain rule, this can be rewritten as
\begin{equation}
	\begin{split}
		\lreggicon = \bigl\|((\nabla\Phi^{AB})(\Phi_\theta^{BA}) + \\
		\varepsilon (\nabla n^{AB}) (\Phi_\theta^{BA})) \cdot (\nabla\Phi^{BA} + \varepsilon \nabla n^{BA}) - \mathbf{I}\bigr\|_F^2\enspace.
		\label{eqn:postchain}
	\end{split}
\end{equation}
Next, we Taylor-expand the term $\nabla \Phi^{AB}(\Phi_\theta^{BA})$, \ie, $\nabla \Phi^{AB} (\Phi^{BA} + \varepsilon n^{BA})$  with respect to $\varepsilon$, yielding
\begin{equation}
	\begin{split}
		& \nabla \Phi^{AB}(\Phi^{BA}_\theta) = \\
		& \nabla \Phi^{AB}(\Phi^{BA}) + \varepsilon \nabla^2 \Phi^{AB}(\Phi^{BA}) n^{BA} + o(\varepsilon)\enspace,
	\end{split}
\end{equation}
where $\nabla^2\Phi^{AB}(\Phi^{BA})n^{BA}$ is the appropriate tensor product.
It is clear that the last term can be dropped in the limit of small $\varepsilon$. Plugging this approximation into \cref{eqn:postchain},
we get
\begin{equation}
	\begin{split}
		\lreggicon \approx \bigl\| & (
		\nabla \Phi^{AB}(\Phi^{BA}) \\
		& + \varepsilon \nabla^2 \Phi^{AB}(\Phi^{BA}) n^{BA} \\
		& + \varepsilon (\nabla n^{AB}) (\Phi_\theta^{BA}))\\
		& \cdot (\nabla\Phi^{BA} + \varepsilon \nabla n^{BA}) - \mathbf{I}\bigr\|_F^2\enspace.
	\end{split}
\end{equation}
Upon distributing terms,
\begin{equation}
	\begin{split}
		\lreggicon \approx \bigl\|
		\nabla \Phi^{AB}(\Phi^{BA}) \cdot \nabla\Phi^{BA}  \\~+
		\varepsilon \nabla^2 \Phi^{AB}(\Phi^{BA}) n^{BA} \cdot \nabla\Phi^{BA} \\~+
		\varepsilon (\nabla n^{AB}) (\Phi_\theta^{BA}) \cdot \nabla\Phi^{BA}  \\~+
		\nabla \Phi^{AB}(\Phi^{BA}) \cdot \varepsilon \nabla n^{BA} \\~+
		\varepsilon \nabla^2 \Phi^{AB}(\Phi^{BA}) n^{BA} \cdot \varepsilon \nabla n^{BA} \\~+
		\varepsilon (\nabla n^{AB}) (\Phi_\theta^{BA}) \cdot \varepsilon \nabla n^{BA}  \\
		- \mathbf{I}\bigr\|_F^2\enspace.
	\end{split}
\end{equation}
The first term equals $\mathbf{I}$ and so cancels with the last term. Further,
by assuming small $\varepsilon$, we drop terms in $\varepsilon ^2$, yielding
\begin{equation}
	\begin{split}
		\lreggicon \approx \bigl\|
		\varepsilon \nabla^2 \Phi^{AB}(\Phi^{BA}) n^{BA} \cdot \nabla\Phi^{BA} \\ +
		\varepsilon (\nabla n^{AB}) (\Phi_\theta^{BA}) \cdot \nabla\Phi^{BA}  \\ +
		\nabla \Phi^{AB}(\Phi^{BA}) \cdot \varepsilon \nabla n^{BA} \bigr\|_F^2\enspace.
	\end{split}
	\label{app:eqn:partialres0}
\end{equation}
Following arguments above regarding the relative magnitudes of $n$ and $\nabla n$, the first term in \cref{app:eqn:partialres0} can be neglected, and also $\varepsilon$ can be factored out to obtain
\begin{equation}
	\begin{split}
		\lreggicon \approx \varepsilon^2 \| & \nabla n^{AB}( \Phi_\theta^{BA}) \nabla \Phi^{BA} \\
		& + \nabla \Phi^{AB}(\Phi^{BA})  \nabla n^{BA}\|_F^2\enspace.
	\end{split}
\end{equation}
\noindent Now, we justify and then use the approximation
\begin{equation}
	\nabla n^{AB}(\Phi_\theta^{BA}) = \nabla n^{AB}(\Phi^{BA}) + O(\varepsilon)\enspace.
\end{equation}
To proceed, we make the following remarks. Inversion of the map preserves a first-order expansion in $\varepsilon$, \ie,
\begin{equation}\label{EqFormulaInverseFirstOrder}
	[\Phi_\theta^{AB}]^{-1} = \Phi^{BA} -\varepsilon \nabla \Phi^{BA}(n^{AB}(\Phi^{BA})) + o(\varepsilon)\enspace,
\end{equation}
which can be checked by composition. A similar first-order expansion holds for the Jacobian determinant, \ie,
\begin{equation}\label{EqJacobianApproximation}
	\operatorname{Det}(\nabla [\Phi_\theta^{AB}]^{-1}) = \operatorname{Det}(\nabla [\Phi^{AB}]^{-1}) + O(\varepsilon)\enspace.
\end{equation}
As a consequence, for a differentiable function (possibly vector valued) $I$, we have
\begin{equation}
	I([\Phi_\theta^{BA}]^{-1}(x)) = I([\Phi^{BA}]^{-1}(x)) + O(\varepsilon)\,,
\end{equation}
by \cref{EqFormulaInverseFirstOrder} and first-order Taylor expansion of $I$. We combine the three above equations in what follows.
Since $\nabla n^{AB}$ is a white noise, this formula is justified in the following sense. For a given vector-valued differentiable function $I$ on the image domain $\Omega$, we have
\begin{equation}
	\resizebox{\columnwidth}{!}{$
			\begin{split}
				& \int_\Omega \nabla n^{AB}((\Phi_\theta^{BA}(x))) \cdot I(x)\, dx \\ &= \int_\Omega \nabla n^{AB}(x) \cdot I([\Phi_\theta^{BA}]^{-1}(x))\operatorname{Det}(\nabla [\Phi_\theta^{BA}]^{-1}(x))\,\mathrm{d}x \\
				& = \int_\Omega \nabla n^{AB}(x) \cdot I([\Phi_\theta^{BA}]^{-1}(x))\operatorname{Det}(\nabla [\Phi^{BA}]^{-1}(x)) \,\mathrm{d}x + O(\varepsilon)\\
				& = \int_\Omega \nabla n^{AB}(\Phi^{BA})\cdot I(x)\,\mathrm{d}x + O(\varepsilon)\enspace,
			\end{split}$
	}
\end{equation}
where, in the first and the last equation, we used the change of variable formula.
This brings us to
\begin{equation}
	\begin{split}
		\lreggicon \approx \varepsilon^2 \| & \nabla n^{AB}( \Phi^{BA}) \nabla \Phi^{BA} \\
		& + \nabla \Phi^{AB}(\Phi^{BA})  \nabla n^{BA}\|_F^2\enspace,
	\end{split}
\end{equation}
which is \cref{EqExpansion2} in the main text.
We now expand the square to get
\begin{equation}
	\begin{split}
		\lreggicon\,{\approx}\,\varepsilon^2\left(\right. \| \nabla n^{AB}( \Phi^{BA}) \nabla \Phi^{BA} \|_F^2 \\ ~+
		\| \nabla \Phi^{AB}(\Phi^{BA})  \nabla n^{BA}\|_F^2 \\~+
		2 \langle  \nabla n^{AB}(\Phi^{BA}) \nabla \Phi^{BA},\left[\nabla \Phi^{AB}\right](\Phi^{BA})\nabla n^{BA} \rangle_{F}                    \left.\right)
	\end{split}
\end{equation}
and an application of the fact that $\Phi^{AB}$ and $\Phi^{BA}$ are inverses of each other gives
\begin{equation}
	\begin{split}
		& \hspace{-0.8cm}\lreggicon \approx \\
		\hspace{0.15cm}\varepsilon^2
		\Bigl(
		& \left\| \nabla n^{AB}(\Phi^{BA}) \nabla \Phi^{BA}\right\|^2_{F}~+ %
		\left\| \left[\nabla \Phi^{BA}\right]^{-1}\nabla n^{BA} \right\|^2_{F} \\
		& \, + 2 \langle  \nabla n^{AB}(\Phi^{BA}) \nabla \Phi^{BA},\left[\nabla \Phi^{BA}\right]^{-1}\nabla n^{BA} \rangle_{F}
		\Bigr)\,.
	\end{split}
	\vspace{-0.2cm}
\end{equation}
When \emph{taking expectation} in \cref{EqTakingExpectation} of the main text, the white noise independence assumption implies that
\begin{equation}
	\mathbb{E}[\nabla n^{BA}_i(y) \nabla n^{AB}_j(x)] = 0
\end{equation}
for all coordinates $i,j$ and $x,y \in \Omega$; this explains that the cross-term vanishes. Thus, we arrive at
\begin{equation*}
	\begin{split}
		\mathbb{E}[\lreggicon] \approx \mathbb{E}\Big[\varepsilon^2
		\Bigl(
		& \left\| \nabla n^{AB}(\Phi^{BA}) \nabla \Phi^{BA}\right\|^2_{F}~+ \\
		& \left\| \left[\nabla \Phi^{BA}\right]^{-1}\nabla n^{BA} \right\|^2_{F} \Bigr)\Big]\,.
	\end{split}
	\vspace{-0.2cm}
\end{equation*}
Now, to further simplify \cref{EqTakingExpectation}, we use the fact that
$\mathbb{E}[\nabla n^{AB}_i(x) \nabla n^{AB}_j(x)] = \delta_{ij}$ where $\delta_{ij} = 1$ if $i=j$ and $\delta_{ij} = 0$ if  $i\neq j$. A direct computation already gives the second term of \cref{Equ:regularization_form} from the main text, \ie,
\begin{equation}
	\begin{split}
		\mathbb{E}[\lreggicon] \approx \mathbb{E}\Big[\varepsilon^2
		\Bigl(
		& \left\| \nabla n^{AB}(\Phi^{BA}) \nabla \Phi^{BA}\right\|^2_{F}~+ \\
		& \left\| \left[\nabla \Phi^{BA}\right]^{-1}\right\|^2_{F} \Bigr)\Big]\,.
	\end{split}
\end{equation}
\vskip0.5ex
In order to obtain the first term of \cref{Equ:regularization_form} from \cref{EqTakingExpectation} in the main text, one needs to use a change of variables $y = \Phi^{AB}(x)$ in space, which results in the appearance of the determinant of the Jacobian matrix, denoted by $\operatorname{Det}(\nabla \Phi^{AB})$, and then similarly use the white noise assumption. The first term of \cref{Equ:regularization_form} has a square root since it is put inside the squared Frobenius norm.
Overall, we arrive at \cref{Equ:regularization_form} from the main text, \ie,
\begin{equation}
	\begin{split}
		\mathbb{E}[\lreggicon]  \approx \varepsilon^2 &
		\Biggl(
		\left\|
		\left[\nabla \Phi^{AB}\right]^{-1} \sqrt{\operatorname{Det}(\nabla\Phi^{AB})}
		\right\|_F^2 \\
		& \hspace{0.23cm}
		+ \left\|
		\left[\nabla\Phi^{BA}\right]^{-1}
		\right\|_F^2\Biggr)\enspace.
	\end{split}
\end{equation}
\vskip1ex
\noindent
\textbf{\texttt{GradICON} and preconditioning.}
Recall that in our notation $\psi(x) = \Phi^{AB}(\Phi^{BA}(x)) - \operatorname{Id}$. The \texttt{ICON} formulation uses $\| \psi \|_{L^2}^2$, whereas the \texttt{GradICON} formulation uses $\| \nabla \psi\|_{L^2}^2$.
Our goal is to understand the effect of this change on the optimization scheme.
The parameters of the neural networks encoding the map from $A, B$ to $\Phi^{AB}$ are optimized to minimize the overall loss but let us focus on the inverse consistency loss. For each pair $A, B$, automatic differentiation computes the gradient of the loss with respect to $\psi$, which is then backpropagated.  Computing the gradient of the \texttt{GradICON} loss can be done by rewriting
\begin{equation}
	\| \nabla \psi\|_{L^2}^2 = -\langle \psi, \Delta \psi \rangle_{L^2}\enspace,
\end{equation}
where $\Delta$ is the Laplacian. Hence, the gradient is $-2\Delta \psi$, which is also called change of metric or \emph{preconditioning}. This gradient has a particularly clear formulation in Fourier space (denoting by $\hat f(\omega)$ the Fourier transform of $f(x)$) since it reads as
\begin{equation}
	\widehat{ \Delta \psi}(\omega) = -|\omega|^2\hat \psi(\omega)
\end{equation}
and has to be compared with the gradient of the \texttt{ICON} loss which is $\hat \psi$. Low frequencies are thus damped in comparison to high frequencies.
For instance, the gradient flows (steepest descent in cont. time) of the two regularizers are
\begin{equation}
	\begin{split}
		\partial_t \psi(x) & = -\psi(x)\enspace\text{and}\\
		\partial_t \psi(x) & = \Delta \psi(x)\enspace.
	\end{split}
\end{equation}
Please note $\psi(x)$ on the top is the $\psi$ in \texttt{ICON} loss while the one on the bottom corresponds to the $\psi$ in \texttt{GradICON} loss.
In Fourier space, these flows become
\begin{equation}
	\begin{split}
		\partial_t\hat{\psi}(\omega) & = -\hat{\psi}(\omega) \enspace\text{and}\\
		\partial_t\hat{\psi}(\omega) & = -|\omega|^2 \hat{\psi}(\omega)\enspace.
	\end{split}
\end{equation}
Here, the main difference is exponential convergence of the first gradient
flow while for the second one (related to \texttt{GradICON}), the rate of
exponential convergence depends on $| \omega|^2$, \ie, faster convergence
occurring for high spatial frequencies $\omega$. We now still need to
understand why such a preconditioning is beneficial for the overall goal of
the diffeomorphic registration problem. From the discussion above, it is clear
that low-frequency perturbations are less penalized than high-frequency
perturbations. The simplest example is the case of constant perturbations
which are not penalized at all by the penalty on the gradient. \vskip0.5ex The
purpose of inverse consistency is to encourage each one of the transformations
$\Phi^{AB}$ and $\Phi^{BA}$ to be a bijective map. However, a relaxation of
this loss (in the sense of a relaxation of a constraint) which still
encourages invertible maps can be beneficial in the context of diffeomorphic
registration. A possible idea consists in relaxing the constraint of $\psi$ to
be a deformation close to identity, while still being diffeomorphic. As a
consequence, if both $\Phi^{AB}(\Phi^{BA})$ and $\Phi^{BA}(\Phi^{AB})$ are
invertible, then both transformation are also invertible. In fact,
constraining these two compositions to be any diffeomorphism also leads to the
same conclusion.

\vskip0.5ex
Further, it is well-known that perturbation of identity by a map with a small Lipschitz constant remains a diffeomorphism. It is the result of the inverse function theorem, explained in more detail below.
Small variations around identity by a Lipschitz map can be written as $\operatorname{Id} + v$ with $v: \mathbb R^d \to \mathbb R^d$ Lipschitz. We want these perturbations of identity to remain diffeomorphic and, in particular, injective (equivalent to asking for no foldings). A sufficient condition to satisfy injectivity is that $\|v(x) - v(y)\| \leq \varepsilon \|x- y\|$ (\ie, $v$ is $\varepsilon$-Lipschitz) with $\varepsilon <1$. When $v$ is $C^1$, the Lipschitz inequality reduces to saying that $\| \nabla v(x)\| \leq \varepsilon$ for every point $x$ in the domain. In fact, as shown by the inverse function theorem, this condition is also sufficient for $\operatorname{Id} + \varepsilon v$ to be a diffeomorphism.  Recall in our case, the deviation to identity is denoted by $\psi$.
In view of this sufficient condition, one would ideally penalize the maximum value of $\|\nabla \psi(x) \|$.
\vskip0.5ex
To sum up, in order to control the invertibility of $\Phi^{AB}(\Phi^{BA}(x))$, it is better to control $\nabla \psi$ rather than $\psi$ itself.
Last, let us show on a concrete example that the \texttt{ICON} regularizer can be more constrained than the \texttt{GradICON} regularizer. To this end, note that \emph{constant} shifts around identity are still invertible maps but \texttt{ICON}  penalizes too large constant shifts, while \texttt{GradICON} does not at all. Also, having small \texttt{ICON} loss does not guarantee invertibility of the resulting maps.
Indeed, there are maps $v$ with a small $L^2$ norm for which the magnitude of the gradient may be larger than $1$, thereby potentially leading to folds.
Penalizing the maximum value of the norm of the gradient of $\psi$ is better suited to guarantee invertibility when the loss is less than $1$. However, in practice, this loss has a lack of differentiability; using an $L^2$ loss is much simpler, more convenient, and retains some of the nice properties mentioned above.

\subsection{Affine data augmentation details}
\label{sec:affine_augmentation}
When we train using affine augmentation, first, we sample a new pair of images from the dataset. Then, we randomly choose whether the image is flipped along each axis: these choices are shared between images in the pair. Finally, independently for each image in the pair, we sample a $3 \times 4$ matrix (with each entry i.i.d. from a standard Gaussian, denoted as $\mathcal{N}^{(3, 4)}(0, 1)$) that represents an affine warp using homogeneous coordinates. This produces an augmented image $\hat{I}$, \ie,
\begin{align*}\hat{I}(\Vec{x}) = I\left(\left(
		\begin{bmatrix}
			u_1 & 0   & 0   & 0 \\
			0   & u_2 & 0   & 0 \\
			0   & 0   & u_3 & 0
		\end{bmatrix}
		{+}\, \gamma \cdot \mathcal{N}^{(3, 4)}(0, 1) \right)
	\begin{bmatrix}
			\Vec{x} \\
			1
		\end{bmatrix}\right)
\end{align*}
where $u_i \sim \text{Uniform}\{\pm 1\}$ and $\gamma=0.05$.
\subsection{Comparison to other regularizers}
\label{sec:appendix_comparing_to_other_regularizers}
\begin{figure*}[b]
	\centering
	\includegraphics[width=.85\linewidth]{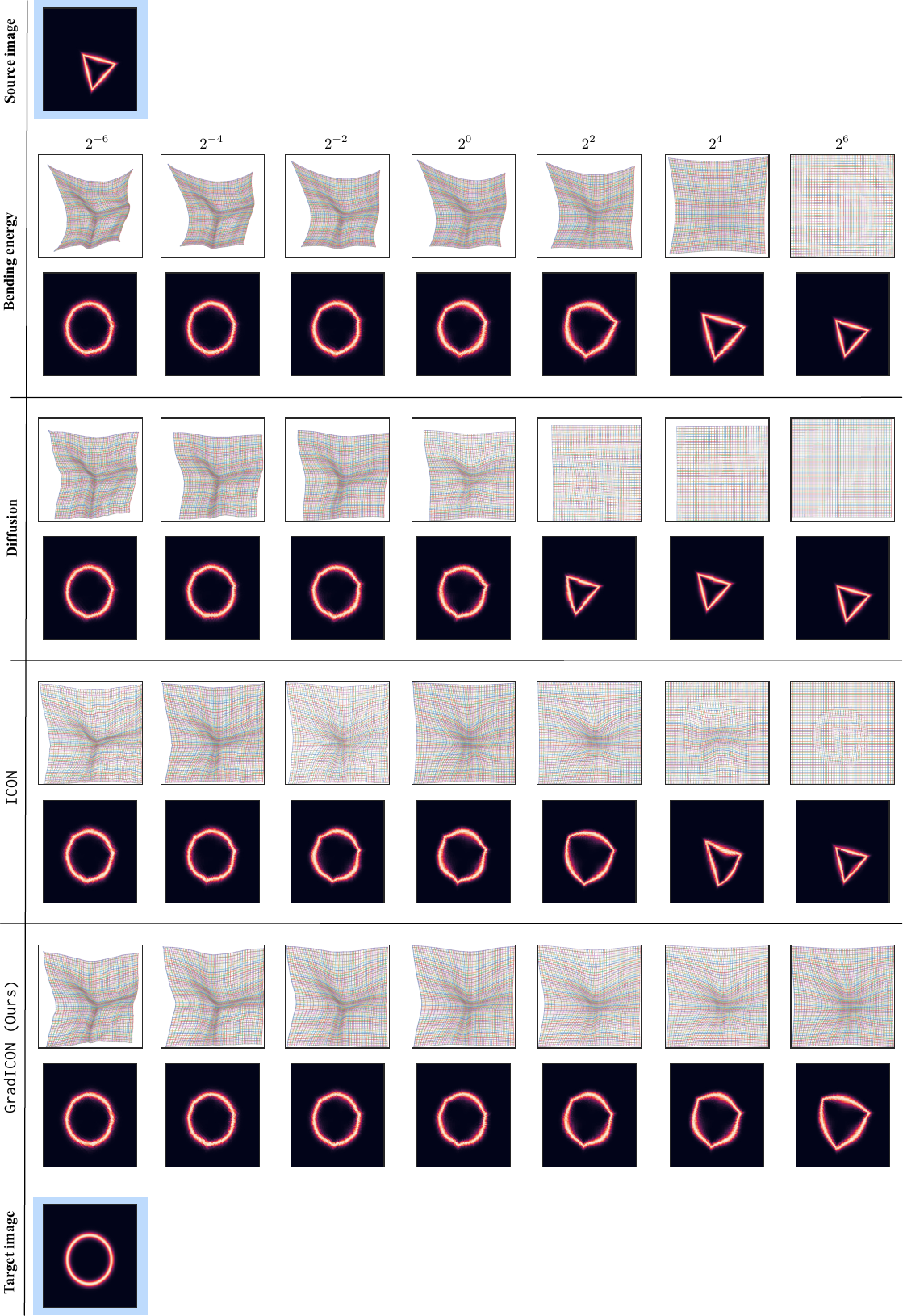}
	\vspace{0.2cm}
	\caption{Illustration of one \emph{warped source image} and the corresponding transformation maps for different regularizers across varying regularization strengths on {\bf Triangles and Circles}. \emph{Best-viewed in color.}}
	\label{Fig:regularizer_comp_example_of_TC}
\end{figure*}

\begin{figure*}[b]
	\centering
	\includegraphics[width=.85\linewidth]{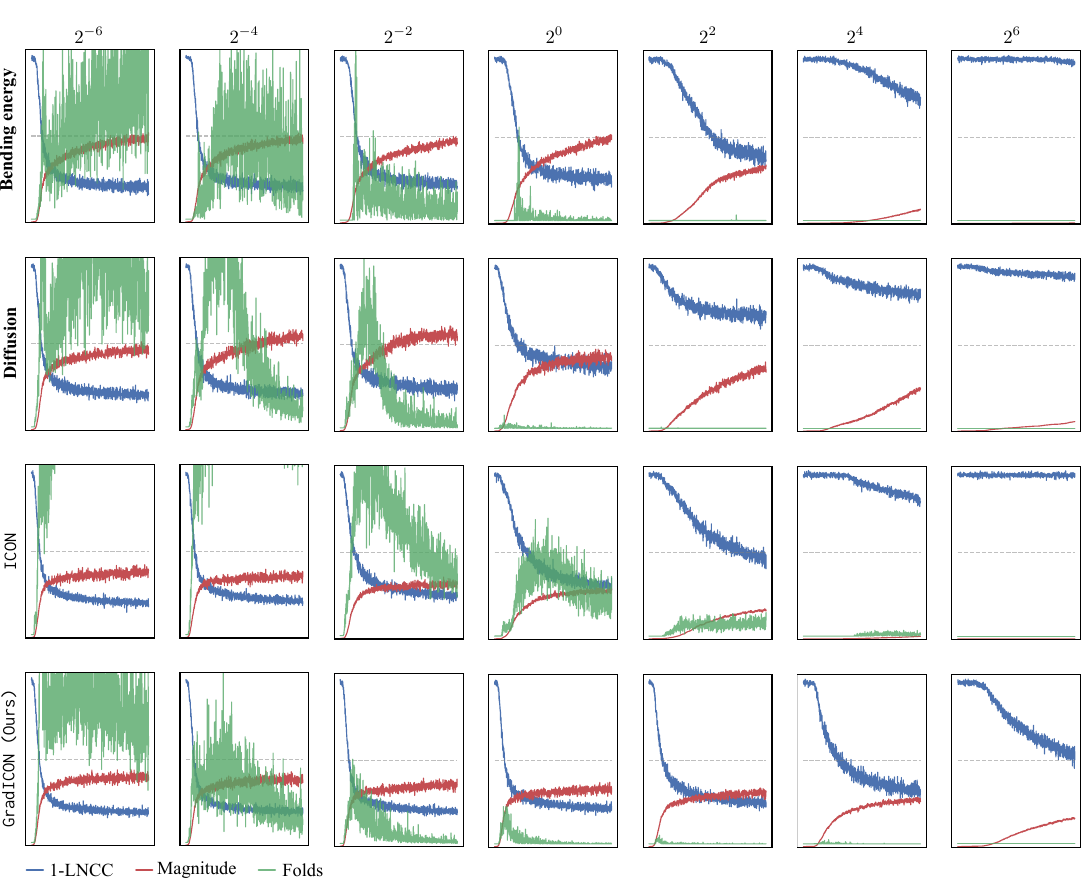}
	\vspace{0.1cm}
	\caption{Illustration of image (dis)similarity (\ie, $1-\text{LNCC}$), the number of folds (Folds), and the mean of the squared $L^2$ norm of the displacement vector field (Magnitude) for different regularizers and across varying regularization strengths on {\bf Triangles and Circles}. \emph{Best-viewed in color.}}
	\label{Fig:regularizer_comp_loss_curve_of_TC}
\end{figure*}

\begin{figure*}[b]
	\centering
	\includegraphics[width=.80\linewidth]{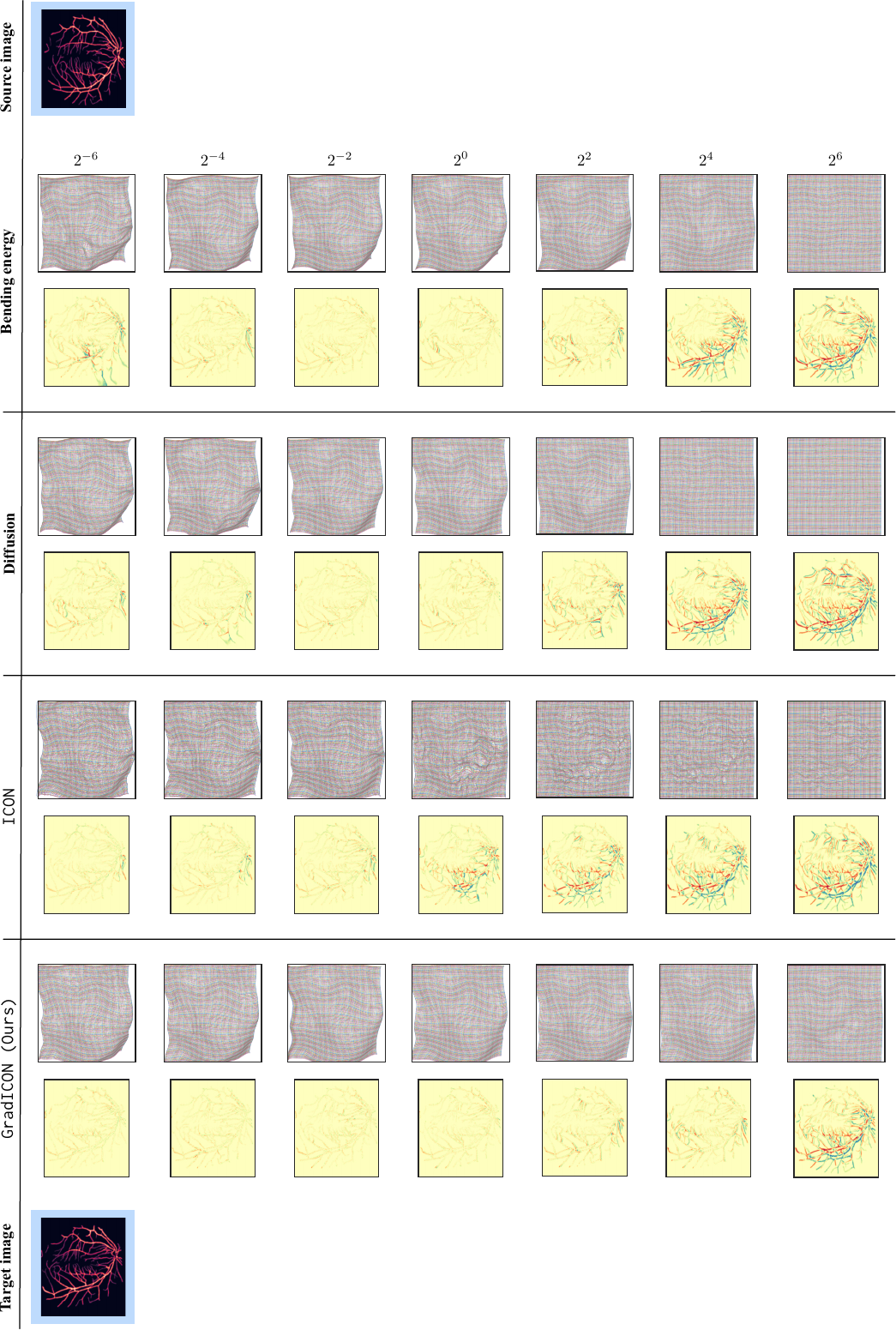}
	\vspace{0.2cm}
	\caption{Illustration of the \emph{residual error} and the corresponding transformation maps between the \emph{warped source image} and the \emph{target image} for different regularizers across varying regularization strengths on {\bf DRIVE}. \emph{Best-viewed in color.}}
	\label{Fig:regularizer_comp_example_of_retina}
\end{figure*}

\begin{figure*}[b]
	\centering
	\includegraphics[width=.85\linewidth]{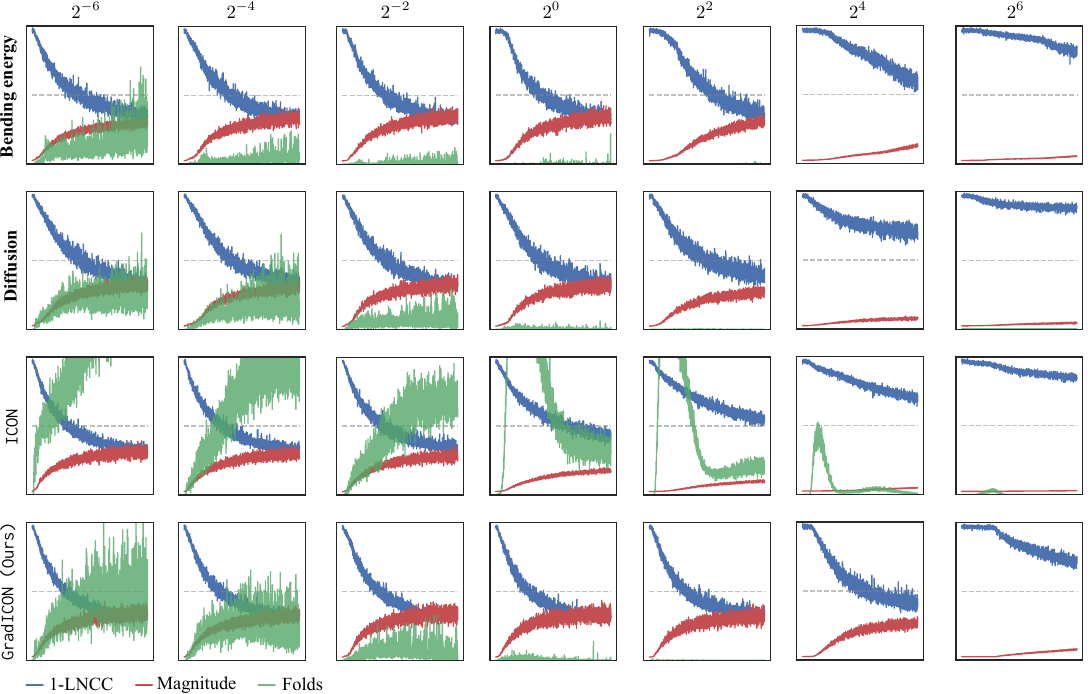}
	\caption{Illustration of image (dis)similarity (\ie, $1-\text{LNCC}$), the number of folds (Folds), and the mean of the squared $L^2$ norm of the displacement vector field (Magnitude) for different regularizers and across varying regularization strengths on {\bf DRIVE}. \emph{Best-viewed in color.}}
	\label{Fig:regularizer_comp_loss_curve_of_retina}
\end{figure*}

In \cref{subsection:regularizercomp} of our main text, we conducted a comparison between different regularizers using displacement vector fields (DVF). However, \cref{Fig:exp_comp_regularizer_with_varying_lambda} only shows \emph{aggregated} results. In  \cref{Fig:regularizer_comp_example_of_TC} and \cref{Fig:regularizer_comp_loss_curve_of_TC}, we present
one registration example from the \textbf{Triangles and Circles} and \textbf{DRIVE} data, respectively. \cref{Fig:regularizer_comp_example_of_retina} and \cref{Fig:regularizer_comp_loss_curve_of_retina} show the image similarity measure, the number of folds and the mean of the squared $L_2$-norm on the displacement vector field plotted over training iterations and with varying regularization weight.

\subsection{Convergence of Gradient Inverse Consistency}
\label{sec:appendix_convergence_experiment}
\begin{figure*}[htp]
	\centering
	\includegraphics[width=0.6\textwidth]{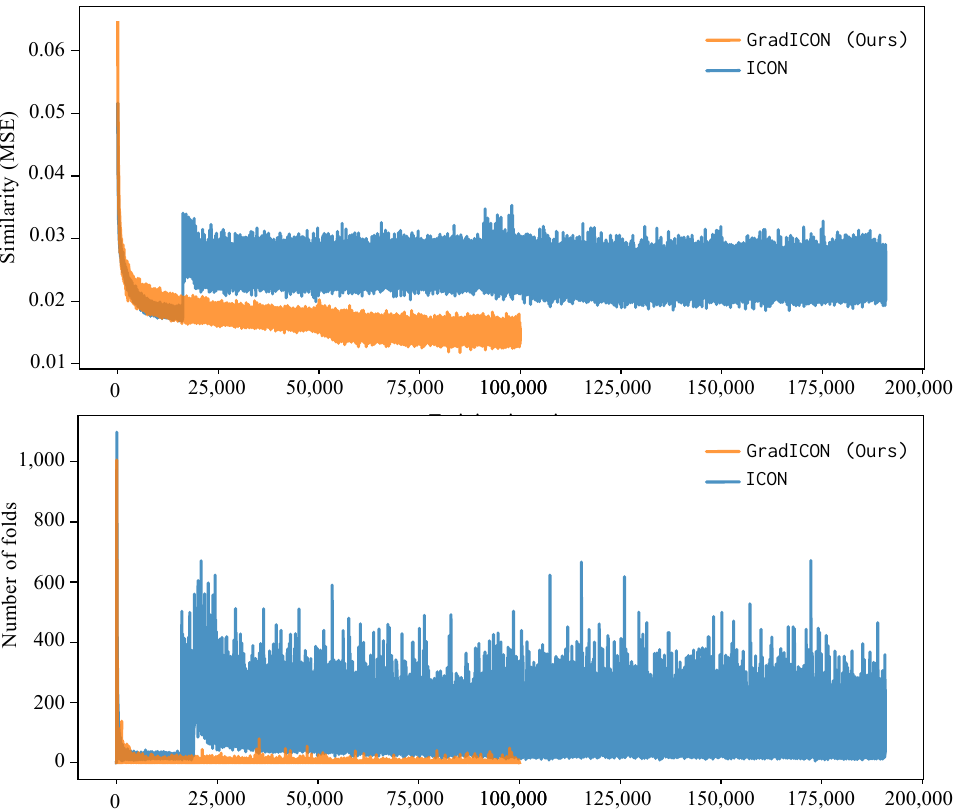}
	\vspace{0.1cm}
	\caption{Image similarity and the number of folds, plotted over training iterations for the \texttt{ICON} and \texttt{GradICON} (MSE, $\lambda=0.2$) entries in the \textbf{OAI} section of \cref{tab:exp_real_datasets_full}. \texttt{ICON}'s parity in similarity loss early in training is illusory, as unlike our approach it is trained progressively, and so during these iterations, it is still being trained at low resolution. This leads to a lower MSE during this phase, as the MSE demands more precise alignment on higher resolution (and hence not low-pass filtered) input images. Once both networks are training at the final resolution, the values are directly comparable. These results demonstrate the faster convergence rate, regularity, and final performance of \texttt{GradICON}.}
	\label{fig:OAI_curves}
\end{figure*}

We conduct the experiment in \cref{sec:exp_convergence_toy_demo} based on our experimental result of \cref{Fig:exp_comp_regularizer_with_varying_lambda}. In particular, we draw a horizontal line in \cref{Fig:exp_comp_regularizer_with_varying_lambda} at $\%|J|=10^{-2}$ and find the closest point to the line on the \texttt{GradICON} and \texttt{ICON} curve. We then plot the loss curves associated with these points in \cref{Fig:exp_convergence_experiment}.
\vskip0.5ex
As can be seen from \cref{Fig:exp_convergence_experiment}, there is a clear convergence speed difference between \texttt{GradICON} and \texttt{ICON} on the two 2D datasets. While a similar study on actual 3D data would be interesting, training 13 models with varying $\lambda$ is computationally challenging.  Nevertheless, to obtain some intuition about convergence speed differences on 3D data, we present in \cref{fig:OAI_curves} the training loss curve, as well as the transform magnitude and the (log) number of folds, of \texttt{GradICON} and \texttt{ICON} corresponding to the \textbf{OAI} dataset in \cref{tab:exp_oai}. \Cref{fig:OAI_curves} clearly supports our claim of  faster convergence of models trained with \texttt{GradICON} regularization over models trained with \texttt{ICON} regularization. Further, it can be seen that \texttt{GradICON} exhibits lower similarity loss and shows larger transform magnitudes because it better captures the large deformations in the \textbf{OAI} dataset.

\subsection{Details for comparisons in Table~\ref{tab:exp_oai}}
\label{sec:appendix_sota_comp}
\begin{table*}[b]
	\centering
	\begin{small}
		\begin{tabular}{lcccccc}
			\toprule
			\bf Method                                            & \bf Trans.              & $\lreg$                               & $\lsim$     & \bf DICE\, $\uparrow$      & $\%|J|\downarrow$ & Reported by                      \\
			\midrule
			\multicolumn{6}{c}{\bf OAI}                                                                                                                                                                                               \\
			Initial                                               &                         &                                       &             & 7.6\resTS                                                                         \\
			\midrule
			Demons~\cite{vercauteren2009diffeomorphic}            & A,DVF                   & Gaussian                              & MSE         & 63.5\resTS                 & 0.0006            & \cite{shen2019networks}          \\
			SyN~\cite{avants2008symmetric}                        & A,VF                    & Gaussian                              & LNCC        & 65.7\resTS                 & 0.0000            & ~\cite{shen2019networks}         \\
			NiftyReg~\cite{modat2010fast}                         & A,B-Spline              & BE                                    & NMI         & 59.7\resTS                 & 0.0000            & ~\cite{shen2019networks}         \\
			NiftyReg~\cite{modat2010fast}                         & A,B-Spline              & BE                                    & LNCC        & {\bf 67.9}\resTS           & 0.0068
			                                                      & \cite{shen2019networks}                                                                                                                                           \\
			vSVF-opt~\cite{shen2019networks}                      & A,vSVF                  & m-Gauss                               & LNCC        & 67.4\resTS                 & 0.0000            & ~\cite{shen2019networks}         \\ \hline
			VM~\cite{balakrishnan2019voxelmorph}                  & SVF                     & Diff.                                 & MSE         & 46.1\resTS                 & 0.0028            & ~\cite{shen2019networks}         \\
			VM~\cite{balakrishnan2019voxelmorph}                  & A,SVF                   & Diff.                                 & MSE         & 66.1\resTS                 & 0.0013            & ~\cite{shen2019networks}         \\ %
			AVSM~\cite{shen2019networks}                          & A,vSVF                  & m-Gauss                               & LNCC        & 68.4\resTS                 & 0.0005            & ~\cite{shen2019networks}         \\ %
			\texttt{ICON}~\cite{greer2021icon}                    & DVF                     & \texttt{ICON}                         & MSE         & 65.1\resTS                 & 0.0040            & $\ast$                           \\ %
			{\textbf{Ours} (MSE, $\lambda{=}0.2$)}                & DVF                     & \cellcolor{black!10}\texttt{GradICON} & MSE         & 69.5\resTS                 & 0.0000            & $\ast$                           \\ %
			{\textbf{Ours} (MSE, $\lambda{=}0.2$, Opt.)}          & DVF                     & \cellcolor{black!10}\texttt{GradICON} & MSE         & 70.5\resTS                 & 0.0001            & $\ast$                           \\ %
			\multirow{2}{*}{\textbf{Ours} \emph{(std. protocol)}} & DVF                     & \cellcolor{black!10}\texttt{GradICON} & LNCC        & 70.1$\dagger$              & 0.0261            & $\ast$                           \\ %
			                                                      & DVF                     & \cellcolor{black!10}\texttt{GradICON} & LNCC        & {\bf 71.2}$\ddagger$       & 0.0042            & $\ast$                           \\
			\midrule
			\multicolumn{6}{c}{\bf HCP}                                                                                                                                                                                               \\
			Initial                                               &                         &                                       &             & 53.4\resTS                 &                                                      \\
			\midrule
			FreeSurfer-Affine~\cite{reuter2010highly}             & A                       & \textemdash                           & TB          & 62.1\resTS                 & 0.0000            & $\ast$                           \\
			SyN~\cite{avants2008symmetric}                        & A,VF                    & Gaussian                              & MI          & \bf{75.8}\resTS           & 0.0000            & $\ast$                           \\ \hline
			sm-shapes~\cite{hoffmann2022synthmorph}               & A,SVF                   & Diff.                                 & DICE        & 79.8\resTS                 & 0.2981            & $\ast$                           \\
			sm-brains~\cite{hoffmann2022synthmorph}               & A,SVF                   & Diff.                                 & DICE        & 78.4\resTS                 & 0.0364            & $\ast$                           \\
			\multirow{2}{*}{\textbf{Ours} \emph{(std. protocol)}} & DVF                     & \cellcolor{black!10}\texttt{GradICON} & LNCC        & 78.7$\dagger$              & 0.0012            & $\ast$                           \\
			                                                      & DVF                     & \cellcolor{black!10}\texttt{GradICON} & LNCC        & \bf{80.5}$\ddagger$        & 0.0004            & $\ast$                           \\
			\midrule
			\multicolumn{6}{c}{\bf DirLab}                                                                                                                                                                                            \\ \midrule
			\bf Method                                            & \bf Trans.              & $\lreg$                               & $\lsim$     & \textbf{mTRE} $\downarrow$ & $\%|J|\downarrow$                                    \\[-2pt]
			                                                      &                         &                                       &             & {\footnotesize [mm]}       &                                                      \\[-2pt]
			Initial                                               &                         &                                       &             & 23.36\resTS                &                                                      \\ \midrule
			SyN~\cite{avants2008symmetric}                        & A,VF                    & Gaussian                              & LNCC        & 1.79\resTS                 & \textemdash       & \cite{hansen2021graphregnet}     \\
			Elastix~\cite{klein2009elastix}                       & A,B-Spline              & BE                                    & MSE         & 1.32\resTS                 & \textemdash       & \cite{hansen2021graphregnet}     \\
			NiftyReg~\cite{modat2010fast}                         & A,B-Spline              & BE                                    & MI          & 2.19\resTS                 & \textemdash       & \cite{hansen2021graphregnet}     \\
			PTVReg~\cite{vishnevskiy2017isotropic}                & DVF                     & TV                                    & LNCC        & 0.96\resTS                 & \textemdash       & ~\cite{vishnevskiy2017isotropic} \\
			RRN~\cite{he2021recursive}                            & DVF                     & TV                                    & LNCC        & \textbf{0.83}\resTS        & \textemdash       & ~\cite{he2021recursive}          \\
			\midrule
			VM~\cite{balakrishnan2019voxelmorph}                  & A,SVF                   & Diff.                                 & NCC         & 9.88\resTS                 & 0                 & $\ast$                           \\
			LapIRN~\cite{mok2020large}                            & SVF                     & Diff.                                 & NCC         & 2.92\resTS                 & 0                 & $\ast$                           \\
			LapIRN~\cite{mok2020large}                            & DVF                     & Diff.                                 & NCC         & 4.24\resTS                 & 0.0105            & $\ast$                           \\
			Hering et al. ~\cite{hering2021cnn}                   & DVF                     & Curv+VCC                              & DICE+KP+NGF & 2.00\resTS                 & 0.0600            & ~\cite{hering2021cnn}            \\
			GraphRegNet~\cite{hansen2021graphregnet}              & DV                      & \textemdash                           & MSE         & 1.34\resTS                 & \textemdash       & ~\cite{hansen2021graphregnet}    \\
			PLOSL~\cite{wang2022PLOSL}                            & DVF                     & Diff.                                 & TVD+VMD     & 3.84\resTS                 & 0                 & ~\cite{wang2022PLOSL}            \\
			PLOSL$_{50}$~\cite{wang2022PLOSL}                     & DVF                     & Diff.                                 & TVD+VMD     & 1.53\resTS                 & 0                 & ~\cite{wang2022PLOSL}            \\
			\texttt{ICON}~\cite{greer2021icon}                    & DVF                     & \texttt{ICON}                         & LNCC        & 7.04\resTS                 & 0.3792            & $\ast$                           \\
			\multirow{2}{*}{\textbf{Ours} \emph{(std. protocol)}} & DVF                     & \cellcolor{black!10}\texttt{GradICON} & LNCC        & 1.93$\dagger$              & 0.0003            & $\ast$                           \\
			                                                      & DVF                     & \cellcolor{black!10}\texttt{GradICON} & LNCC        & \textbf{1.31}$\ddagger$    & 0.0002            & $\ast$                           \\
			\bottomrule\vspace{-5mm}
		\end{tabular}
		\caption{Full comparison on \textbf{OAI}, \textbf{HCP} and \textbf{DirLab}. $\dagger$ and $\ddagger$ indicate results from our standard training protocol, without ($\dagger$) and with ($\ddagger$) instance optimization (\cref{subsection:training}). Only when \texttt{GradICON} is trained with MSE, we set $\lambda=0.2$. \emph{Top} and \emph{bottom} table parts denote non-learning and learning-based methods, resp. For \textbf{DirLab}, results are shown in the common \emph{inspiration$\rightarrow$expiration} direction. Results marked with $^\ast$ are reported by us using code from the official repository. \underline{A}: affine pre-registration, \underline{BE}: bending energy, \underline{MI}: mutual information, \underline{DV}: displacement vector of sparse key points, \underline{TV}: total variation, \underline{Curv}: curvature regularizer, \underline{VCC}: volume change control, \underline{NGF}: normalized gradient flow, \underline{TVD}: sum of squared tissue volume difference, \underline{VMD}: sum of squared vesselness measure difference, \underline{Diff}: diffusion, \underline{VF}: velocity field, \underline{SVF}: stationary VF, \underline{DVF}: displacement vector field. $\underline{\text{PLOSL}_{50}}$: 50 iterations of instance optimization with PLOSL.}
		\label{tab:exp_real_datasets_full}
	\end{small}
\end{table*}
In this section, we 1) present Table~\ref{tab:exp_real_datasets_full}, which is a complete version of Table~\ref{tab:exp_oai}, and 2) describe the experimental details for comparisons in Table~\ref{tab:exp_oai}.
\vskip1ex
\noindent
\textbf{sm-shapes and sm-brains.}
We evaluate the SynthMorph~\cite{hoffmann2022synthmorph} model with pre-trained weights from its official repository\footnote{\label{fn:voxelmorph}\url{https://github.com/voxelmorph/voxelmorph}} on the same \textbf{HCP} test set we use for \texttt{GradICON} and follow the suggested testing protocol from the repository. We first run \textbf{FreeSurfer-Affine} (see the following Freesurfer-Affine paragraph) to align the source and target image to the reference image provided in the repository. Then, we run \emph{SynthMorph-shapes} (\texttt{sm-shapes}) and \emph{SynthMorph-brains} (\texttt{sm-brains}) models to obtain the deformation between the pre-aligned source and target images. To compute the final transformation field used to warp the original source label map to the target label map, we first generate an identity map in the target image space and transform it via the target-to-reference affine matrix. Then, we compose the transformed map with the deformation field computed by SynthMorph. Lastly, we transform the composed deformation field via the reference-to-source affine matrix (obtained by inverting the source-to-reference affine matrix). Eventually, we use the final computed deformation field to warp the original source label map and then compute the DICE between the warped label map and the target label map in the \emph{original} target space.
\vskip1ex
\noindent
\textbf{FreeSurfer-Affine.}
We report the affine pre-alignment result from our SynthMorph experiment and label it as FreeSurfer-Affine~\cite{reuter2010highly} in Table~\ref{tab:exp_oai}. FreeSurfer is run with the configuration recommended in the SynthMorph repository. For evaluation purposes, we compose the target-to-reference affine matrix and reference-to-source affine matrix the same way as we did in the SynthMorph experiment except that we skip the step to compose the deformation field computed by SynthMorph. Essentially, we simply assume that the non-parametric part that would have been obtained by SynthMorph is set to the identity transform thereby only leaving the affine registrations. This experiment differs from directly obtaining an affine transformation between the source and the target spaces as instead we go through the template space and compute two affine transformations. However, this choice of affine transform composition allows a more direct assessment of the improvements obtained by SynthMorph.
\vskip1ex
\noindent
\textbf{\texttt{ICON}.}
We follow a similar design as described in \cite{greer2021icon} and, in particular, adopt the \texttt{tallUNet2} architecture as the backbone network. Specifically, a composition of two such UNet's is initially trained on half-resolution image pairs. This network is then composed with a third UNet, trained on full-resolution image pairs.
\vskip1ex
\noindent
\textbf{VoxelMorph.}
We use the official code from the VoxelMorph repository\footref{fn:voxelmorph} %
and train on \textbf{COPDGene}. As VoxelMorph requires pre-registration, we train another neural network for affine pre-registration. Table~\ref{tab:exp_vm_affine_preregistration} shows the registration accuracy of this affine registration network.
For VoxelMorph, we use NCC as the similarity measure, set the learning rate to 1e-3, and the regularizer weight to 5. We keep all other settings at the provided default values. %
Since the official code provides the inverse transformation, the bi-directional registration result is obtained with the forward map and its inverse map is computed by the official code.
\begin{table*}[htp]
	\centering
	\begin{small}
		\begin{tabular}{ccc}
			\toprule
			       & \bf mTRE\, $\downarrow$ & \bf DICE\, $\uparrow$ \\ \midrule
			Affine & 13.715                  & 80.23                 \\\bottomrule
		\end{tabular}
	\end{small}
	\caption{Registration performance measures of the pre-registration \emph{affine} network for our VoxelMorph comparison on \textbf{DirLab}.}
	\label{tab:exp_vm_affine_preregistration}
\end{table*}
\vskip1ex
\noindent
\textbf{LapIRN.}
We obtain the network from the official repository\footnote{\url{https://github.com/cwmok/LapIRN}} of LapIRN and train it on \textbf{COPDGene}. In particular, we train using the training script provided by the official repository with hyperparameter tuning for {\bf COPDGene} data. We switched from LNCC to NCC because we observed unstable training with the LNCC implementation provided in the official LapIRN repository. For each resolution, we adjust the learning rate and $\lambda$ to assure that the training converges. Table~\ref{tab:exp_lapirn_parameters} provides the hyperparameters we used to obtain the results in Table~\ref{tab:exp_oai}. We randomly swap the source and target images during training so that the trained network can work for bi-directional registration.
\begin{table*}[htp]
	\centering
	\begin{small}
		\begin{tabular}{ccccc}
			\toprule
			\multicolumn{1}{c}{\multirow{2}{*}{Resolution}} & \multicolumn{2}{c}{LapIRN (disp)} & \multicolumn{2}{c}{LapIRN (sVF)}                                  \\
			\cmidrule{2-5}
			\multicolumn{1}{c}{}                            & \multicolumn{1}{c}{\bf LR}        & \multicolumn{1}{c}{\bf reg. weight} & \bf LR    & \bf reg. weight \\ \midrule
			1                                               & $1e^{-4}$                         & 0.1                                 & $1e^{-4}$ & 0.1             \\
			$\sfrac{1}{2}$                                  & $5e^{-5}$                         & 0.1                                 & $1e^{-4}$ & 0.1             \\
			$\sfrac{1}{4}$                                  & $1e^{-5}$                         & 1                                   & $5e^{-5}$ & 1               \\ \bottomrule
		\end{tabular}
		\caption{\label{tab:exp_lapirn_parameters} Learning rate (LR) and regularization weight (reg. weight) hyperparameters of LapIRN per resolution.}
	\end{small}
\end{table*}

\subsection{Model statistics}\label{sec:model_statistics}
We compute model statistics regarding the number of parameters, peak memory
use, FLOPs, and inference time using built-in
PyTorch\footnote{\url{https://pytorch.org/}} functions and the
\texttt{thop}\footnote{\url{https://github.com/Lyken17/pytorch-OpCounter}}
package. This experiment is conducted using an NVIDIA GeForce RTX 3090 GPU
with a batch size of 1 and randomly generated image pairs of size
$175\times175\times175$. We run the model 10 times and take the average of the
elapsed time as the final measurement. In addition, the peak GPU memory usage
is reported for each model. Table~\ref{tab:model_statistics_lung} and
Table~\ref{tab:model_statistics_unet} list the statistics of models evaluated
in Table~\ref{tab:exp_oai} and the UNet used in our ablation study of
Sec.~\ref{subsection:ablationstudy}. Working on 2-D convolutional networks
builds a strong intuition that if, in a downsampling step, we double the
number of channels and cut in half the resolution, the amount of computation
stays roughly constant. This intuition is not correct for 3-D networks. In
fact, for 3-D networks, doubling the number of channels and halving the
resolution cuts the amount of computation by 1/2. As a result, large channel
counts deep in the network are, from a computation time and VRAM perspective,
free. The UNet from \texttt{ICON} approach~\cite{greer2021icon} takes
advantage of this effect to boost performance using a large parameter count
while reducing runtime and VRAM usage compared to the standard VoxelMorph
channel counts. We used the same approach for our registration networks using
\texttt{GradICON}. Note that Table~\ref{tab:model_statistics_lung} illustrates
that even though \texttt{ICON} and \texttt{GradICON} use about 50 times more
parameters than LapIRN and roughly 150 times more parameters than VoxelMorph,
memory consumption and inference times are in fact lower.
\begin{table*}[h!]
	\begin{small}
		\centering
		\vspace{\baselineskip}
		\begin{tabular}{lcccccc}\toprule
                                              & \multirow{2}{*}{\bf \#Params} & \multicolumn{3}{c}{Inference} & \multicolumn{2}{c}{Train} \\ \cmidrule{3-7}
			                                 &  & \textbf{Peak Mem.}~(MB) & \textbf{FLOPs} & \textbf{Time}~(ms) & \textbf{Peak Mem.}~(MB) & \textbf{Time}~(ms) \\ \midrule
			VM (SVF)                         & 327,331      & 4548                    & 397.878G       & 190.10  & \textemdash       & \textemdash            \\
			LapIRN (SVF)                     & 923,748      & 5578                    & 652.310G       & 253.87  & \textemdash       & \textemdash            \\
			LapIRN (DV)                      & 923.748      & 5576                    & 652.310G       & 235.30  & \textemdash       & \textemdash            \\
			\texttt{ICON}                    & 53,010,687   & 2918                    & 678.513G       & 96.36   & 8082              & 573.57                 \\
			\texttt{GradICON-Stage1}         & 53,010,687   & 2934                    & 618.592G       & 88.24   & 9384              & 727.77                 \\
			\texttt{GradICON-Stage1\&Stage2} & 70,680,916   & 3122                    & 1.159T         & 160.59  & 13482                & 1162.54                     \\
			\bottomrule
		\end{tabular}
		\caption{Model statistics at inference (test) time. \underline{G} denotes gigaFLOPS, \underline{T} denotes teraFLOPS.}
		\label{tab:model_statistics_lung}
	\end{small}
\end{table*}
\begin{table*}[htp]
	\centering
	\begin{small}
		\vspace{\baselineskip}
		\begin{tabular}{lccc}\toprule
			                                            & \bf \#Params & \textbf{Peak Mem.}~(MB) & \textbf{FLOPs} \\ \midrule
			UNet from~\cite{balakrishnan2019voxelmorph} & 327,331      & 4182.0                  & 397.878G       \\
			UNet from~\cite{greer2021icon}              & 17,670,229   & 2244                    & 540.084G       \\
			\bottomrule
		\end{tabular}
		\caption{Model statistics of the UNets used in our ablation study. \underline{G} denotes gigaFLOPS.}
		\label{tab:model_statistics_unet}
	\end{small}
\end{table*}

\section{Visualizations}
\label{appendix:visualizations}
\begin{figure*}
	\includegraphics[width=\textwidth]{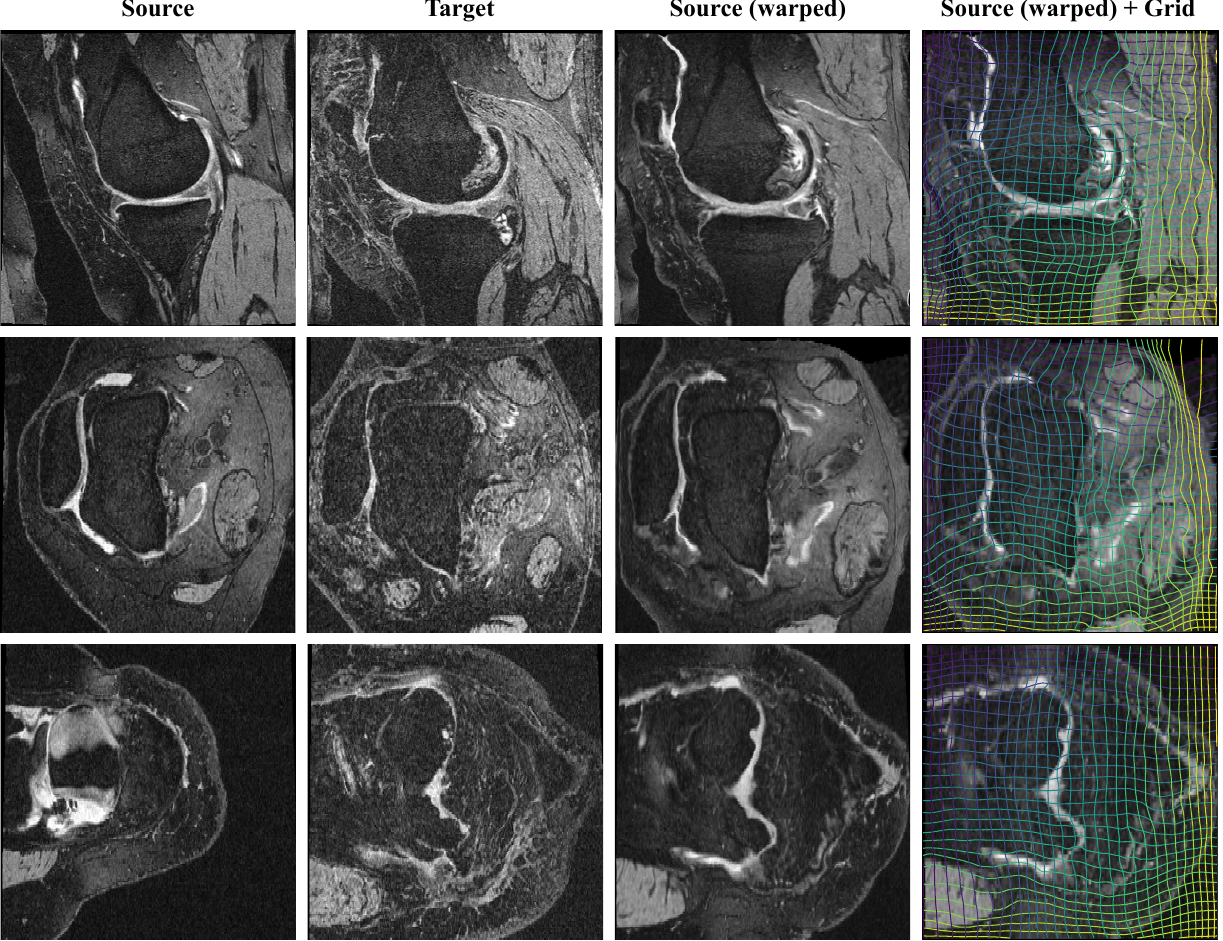}
	\caption{Example registration case \textbf{A} (from test set instances) performed using \texttt{GradICON} and our standard training protocol ($\dagger$) w/o instance optimization on the \textbf{OAI} dataset. \emph{Best-viewed in color.}\label{Fig:knee_example_OAI0}}
\end{figure*}

\begin{figure*}
	\includegraphics[width=\textwidth]{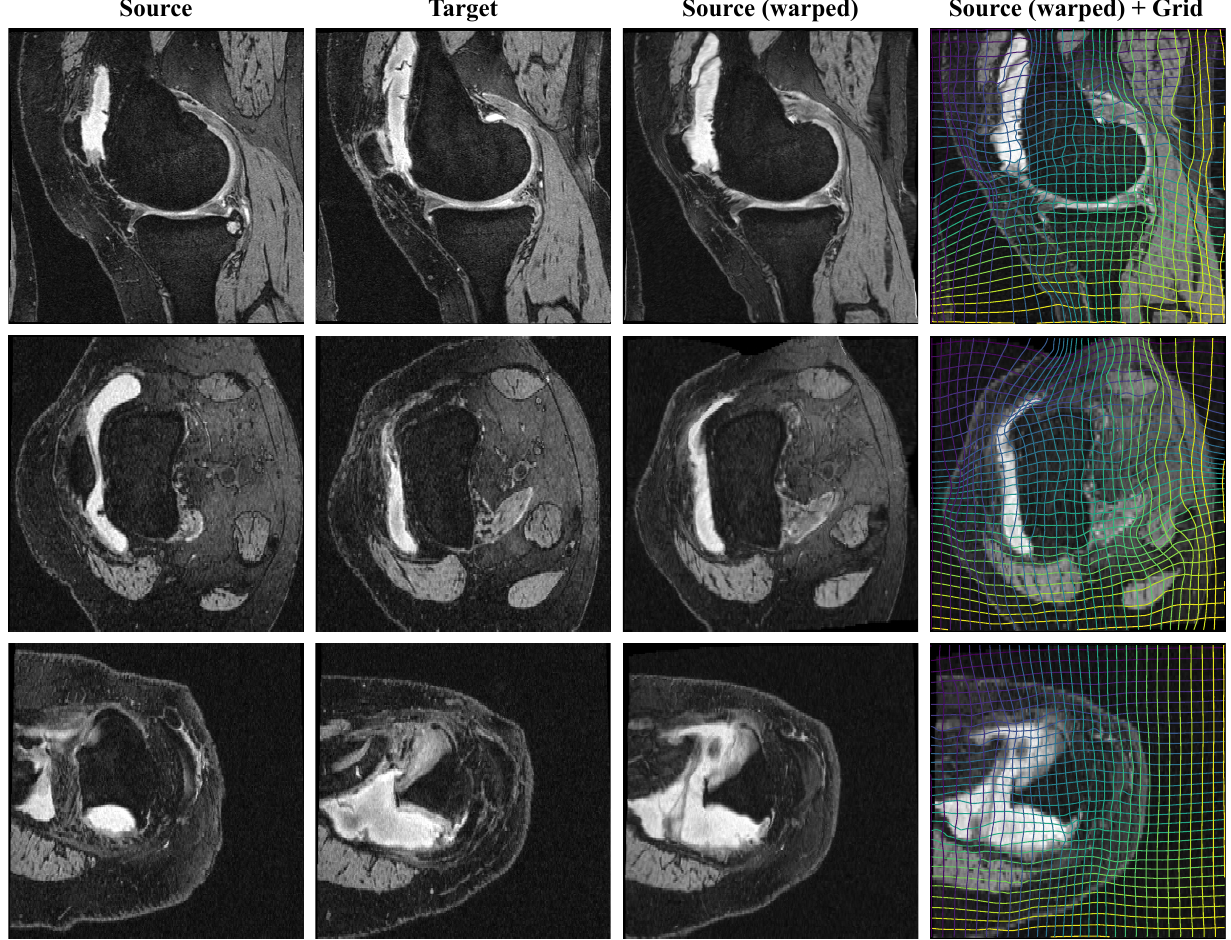}
	\caption{Example registration case \textbf{B} (from test set instances) performed using \texttt{GradICON} and our standard training protocol ($\dagger$) w/o instance optimization on the \textbf{OAI} dataset. \emph{Best-viewed in color.}\label{Fig:knee_example_OAI1}}
\end{figure*}

\begin{figure*}
	\includegraphics[width=\textwidth]{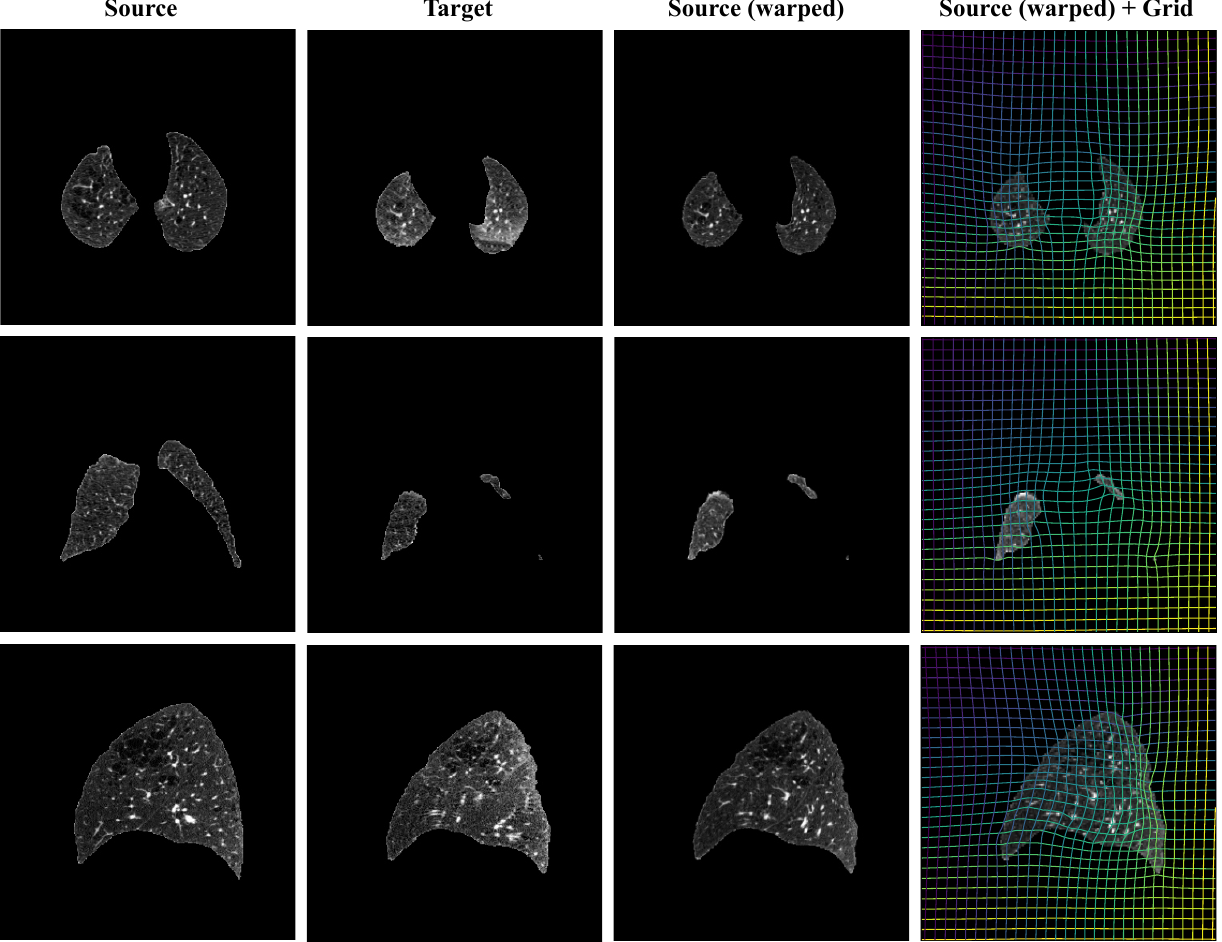}
	\caption{Example registrations case \textbf{A} performed using \texttt{GradICON} and our standard training protocol ($\dagger$) w/o instance optimization on the \textbf{Dirlab} dataset. \emph{Best-viewed in color.}\label{Fig:lung_example_DirLab1}}
\end{figure*}

\begin{figure*}
	\includegraphics[width=\textwidth]{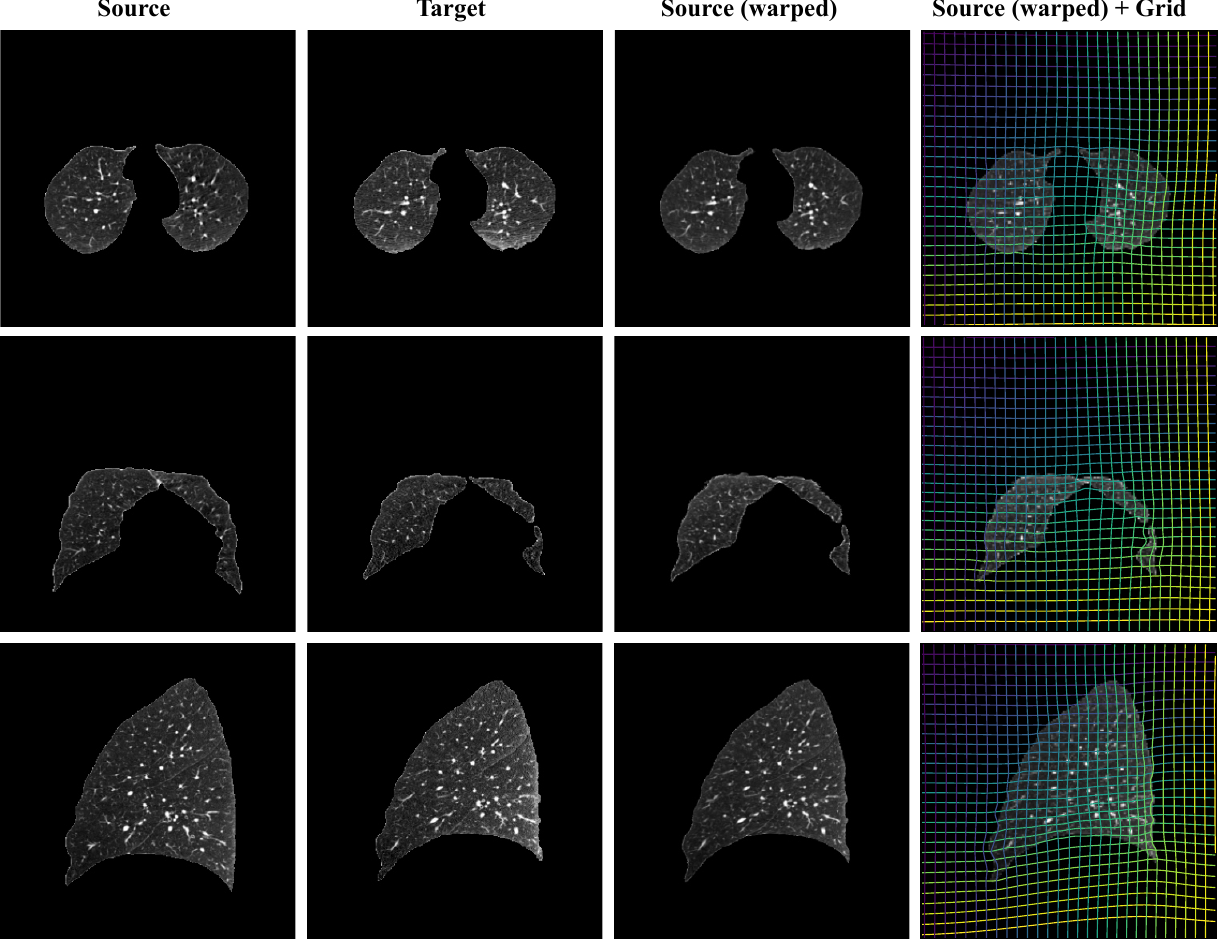}
	\caption{Example registrations case \textbf{B} performed using \texttt{GradICON} and our standard training protocol ($\dagger$) w/o instance optimization on the \textbf{Dirlab} dataset. \emph{Best-viewed in color.}\label{Fig:lung_example_DirLab2}}
\end{figure*}

\begin{figure*}
	\includegraphics[width=\textwidth]{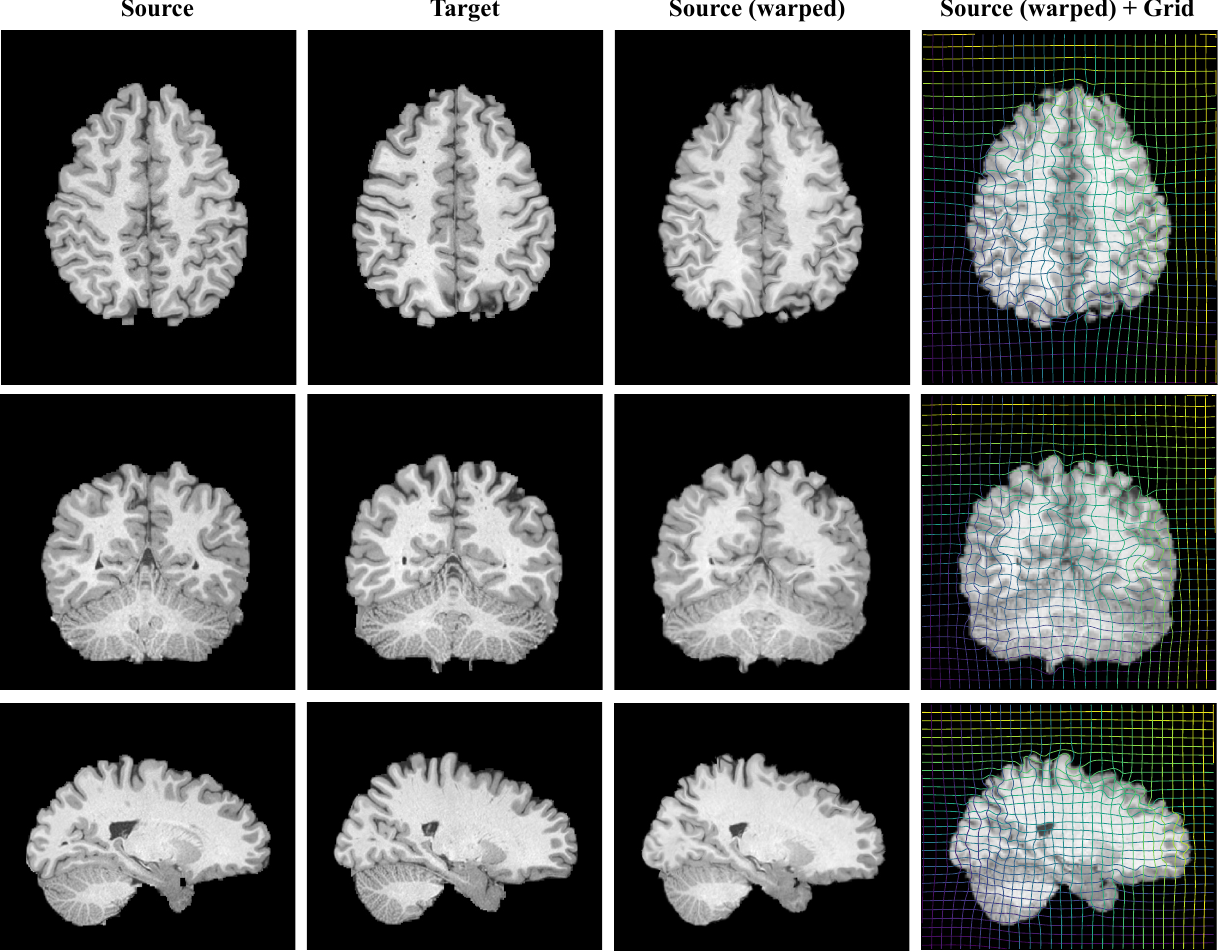}
	\caption{\label{Fig:brain_example_case1} Example registration case \textbf{A} (from test set instances) performed using \texttt{GradICON} and our standard training protocol ($\dagger$) w/o instance optimization on the \textbf{HCP} dataset. \emph{Best-viewed in color.}}
\end{figure*}

\begin{figure*}
	\includegraphics[width=\textwidth]{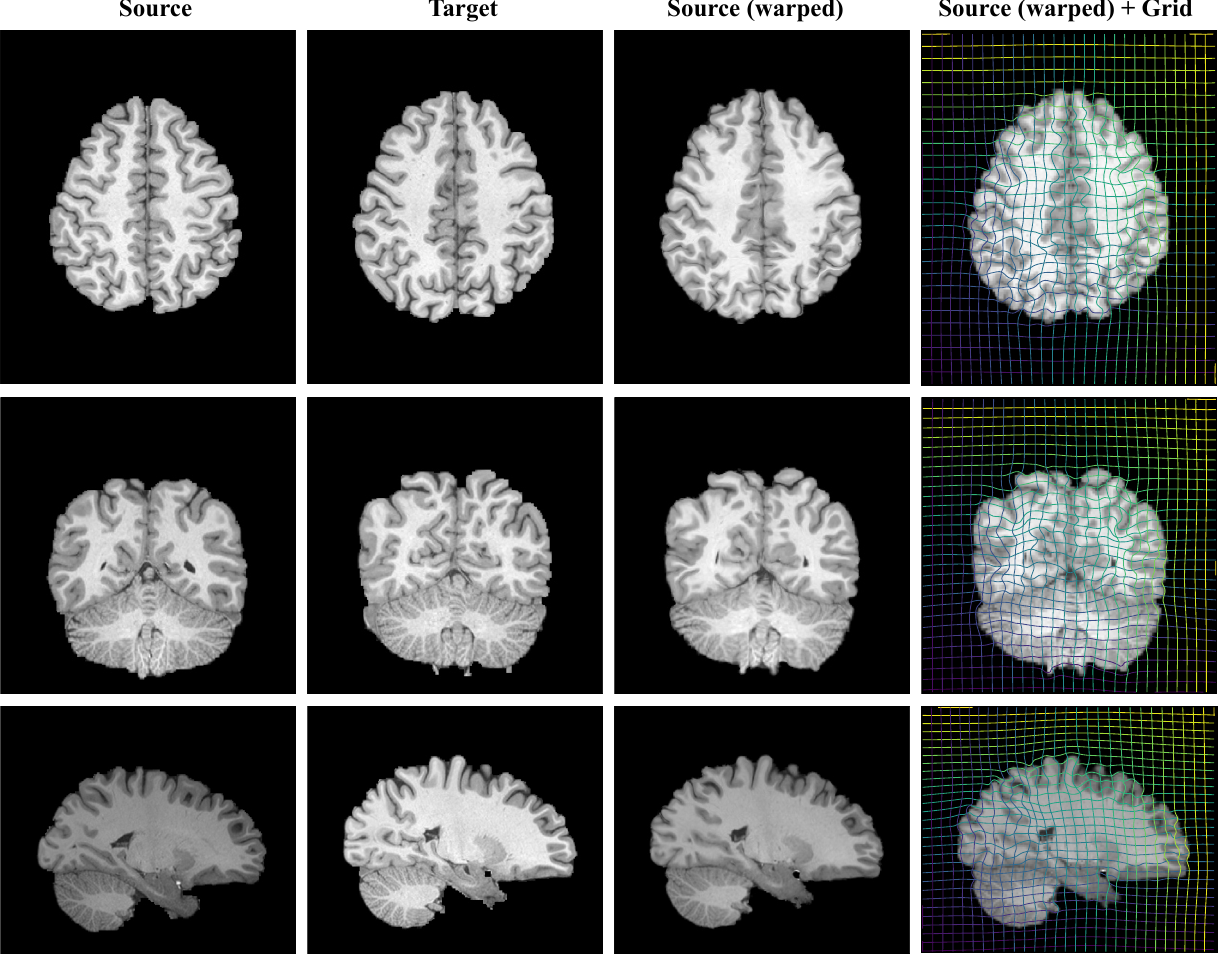}
	\caption{\label{Fig:brain_example_case2} Example registration case \textbf{B} (from test set instances) performed using \texttt{GradICON} and our standard training protocol ($\dagger$) w/o instance optimization on the \textbf{HCP} dataset. \emph{Best-viewed in color.}}
\end{figure*}

In Fig.~\ref{Fig:knee_example_OAI0} and Fig.~\ref{Fig:knee_example_OAI1}, we show two example
registration cases from \textbf{OAI}, Fig.~\ref{Fig:lung_example_DirLab1} and Fig.~\ref{Fig:lung_example_DirLab2} show two example registration cases on \textbf{DirLab}, and Fig.~\ref{Fig:brain_example_case1} and Fig.~\ref{Fig:brain_example_case2} show two example
registration cases on \textbf{HCP}.

\begin{figure*}
\centering
    \includegraphics[width=.7\textwidth]{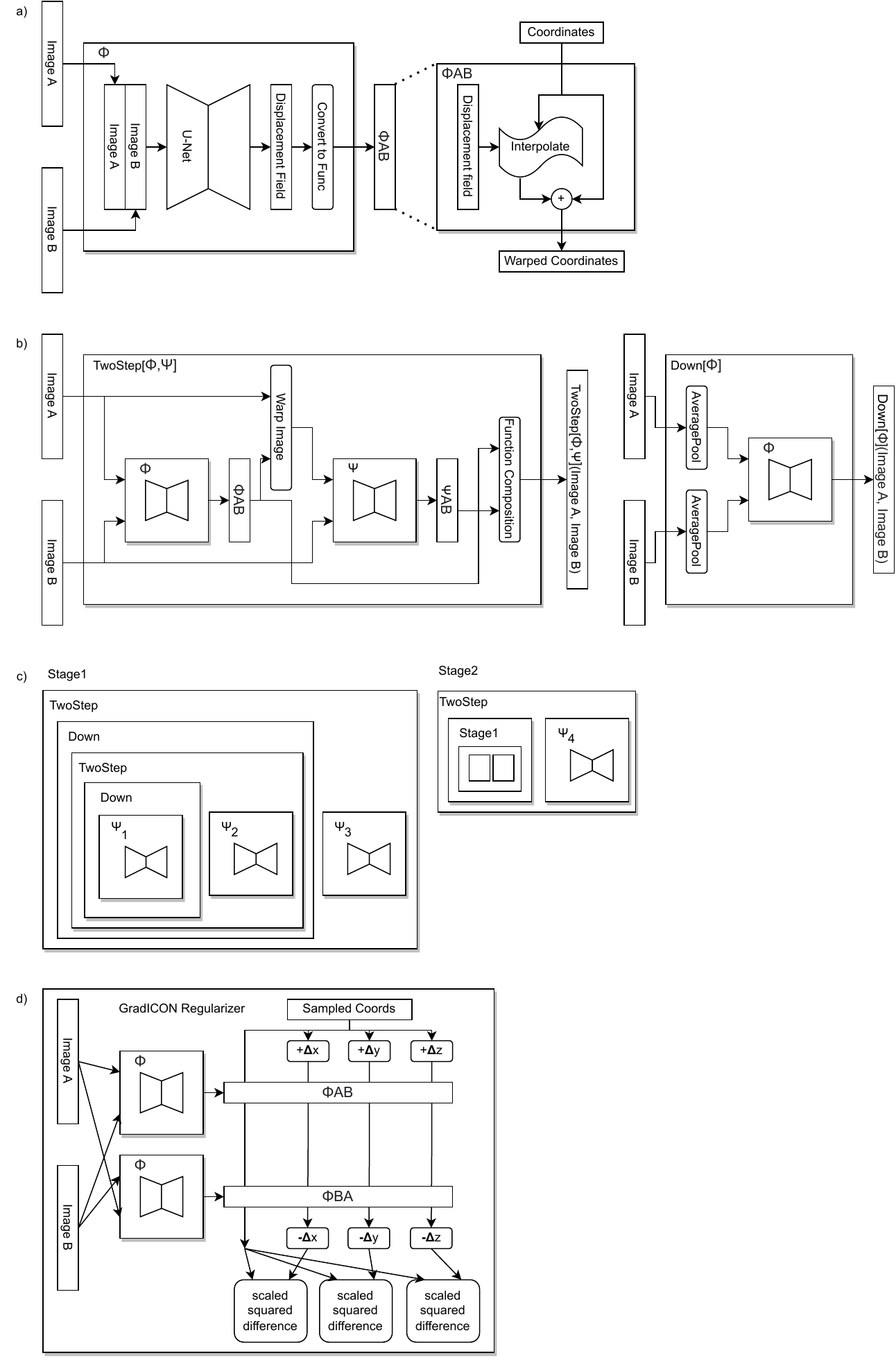}
    \caption{Our approach is most succinctly described using equations, as done in \cref{sec:network_architecture}
, but we also desire to respect the convention that neural network papers include a representation of the network as a block diagram. Our "atomic," or simplest component registration network is a U-Net outputting a deformation (a). $\Phi^{AB}$, the output of this component, is a \emph{python function} that may be called on a tensor of coordinates. Components can be combined using the \texttt{TwoStep} and \texttt{Down} operators (b). The 'function composition' block in this row is implemented by the python code \texttt{lambda coords: phi\_AB(psi\_AB(coords))} , which is pleasing enough to justify our decision to represent deformations as functions. These parts are combined into the Stage1 and Stage2 networks we use for our general purpose registration approach (c). Finally, this network is regularized by a finite difference approximation of the gradient of the inverse consistency error (d)}
\label{Fig:network_block_diagram}
\end{figure*}

In Fig.~\ref{Fig:network_block_diagram} we show a block diagram of the network structure described in Sec.~\ref{sec:network_architecture}, that is more detailed.

\section{Potential negative societal impacts}
\label{Sec:social_impact}
Image registration results might not be accurate or might even fail for certain image pairs in practice. Hence, careful quality control of the results should be performed when registrations are used for decision-support systems in a medical context.
\end{document}